\documentclass[10pt,journal,letterpaper,compsoc]{IEEEtran}
\usepackage{amsmath,amsfonts,amssymb,epsfig,verbatim,color,times}
\usepackage{graphicx}
\usepackage{multicol,multirow}
\usepackage{booktabs}
\usepackage{epsfig,graphicx,subfig,pdfpages,rotating,float}
\usepackage{algorithm}
\usepackage{algpseudocode}
\usepackage{url}
\usepackage{bm,array,mathtools}
\usepackage{xcolor,colortbl}
\usepackage{textcomp}

\newcommand{\argmin}{\operatornamewithlimits{\arg\,\min}}

\def\@onedot{\ifx\@let@token.\else.\null\fi\xspace}

\renewcommand{\v}[1]{\ensuremath{\mathbf{#1}}}
\newcommand{\etal}{\mbox{\emph{et al.\ }}}

\DeclareMathAlphabet{\mathpzc}{OT1}{pzc}{m}{it}

\begin{document}

\title{Re-identification of Humans in Crowds using \\Personal, Social and Environmental Constraints}

\author{Shayan~Modiri~Assari,~\IEEEmembership{Member,~IEEE,}
        Haroon~Idrees,~\IEEEmembership{Member,~IEEE,}
        and~Mubarak~Shah,~\IEEEmembership{Fellow,~IEEE}
\IEEEcompsocitemizethanks{\IEEEcompsocthanksitem S. Modiri Assari, H. Idrees and M. Shah are with the Center for Research in Computer Vision (CRCV), University of Central Florida, Orlando, FL, 32816. E-mail: \{smodiri,haroon,shah\}@eecs.ucf.edu}}

\markboth{IEEE TRANSACTIONS ON PATTERN ANALYSIS AND MACHINE INTELLIGENCE}%
{Shell \MakeLowercase{\textit{et al.}}: Bare Demo of IEEEtran.cls for Computer Society Journals}

\IEEEcompsoctitleabstractindextext{
\begin{abstract}

This paper addresses the problem of human re-identification across non-overlapping cameras in crowds.
Re-identification in crowded scenes is a challenging problem due to large number of people and frequent occlusions, coupled with changes in their appearance due to different properties and exposure of cameras. To solve this problem, we model multiple Personal, Social and Environmental (PSE) constraints on human motion across cameras. The personal constraints include appearance and preferred speed of each individual assumed to be similar across the non-overlapping cameras. The social influences (constraints) are quadratic in nature, i.e. occur between pairs of individuals, and modeled through grouping and collision avoidance. Finally, the environmental constraints capture the transition probabilities between gates (entrances / exits) in different cameras, defined as multi-modal distributions of transition time and destination between all pairs of gates. We incorporate these constraints into an energy minimization framework for solving human re-identification. Assigning $1-1$ correspondence while modeling PSE constraints is NP-hard. We present a stochastic local search algorithm to restrict the search space of hypotheses, and obtain $1-1$ solution in the presence of linear and quadratic PSE constraints. Moreover, we present an alternate optimization using Frank-Wolfe algorithm that solves the convex approximation of the objective function with linear relaxation on binary variables, and yields an order of magnitude speed up over stochastic local search with minor drop in performance. We evaluate our approach using Cumulative Matching Curves as well $1-1$ assignment on several thousand frames of Grand Central, PRID and DukeMTMC datasets, and obtain significantly better results compared to existing re-identification methods.

\end{abstract}

}

\maketitle

\IEEEdisplaynotcompsoctitleabstractindextext \IEEEpeerreviewmaketitle

\section{Introduction}

\begin{figure*}[t]
\centering
\includegraphics[width=1\textwidth]{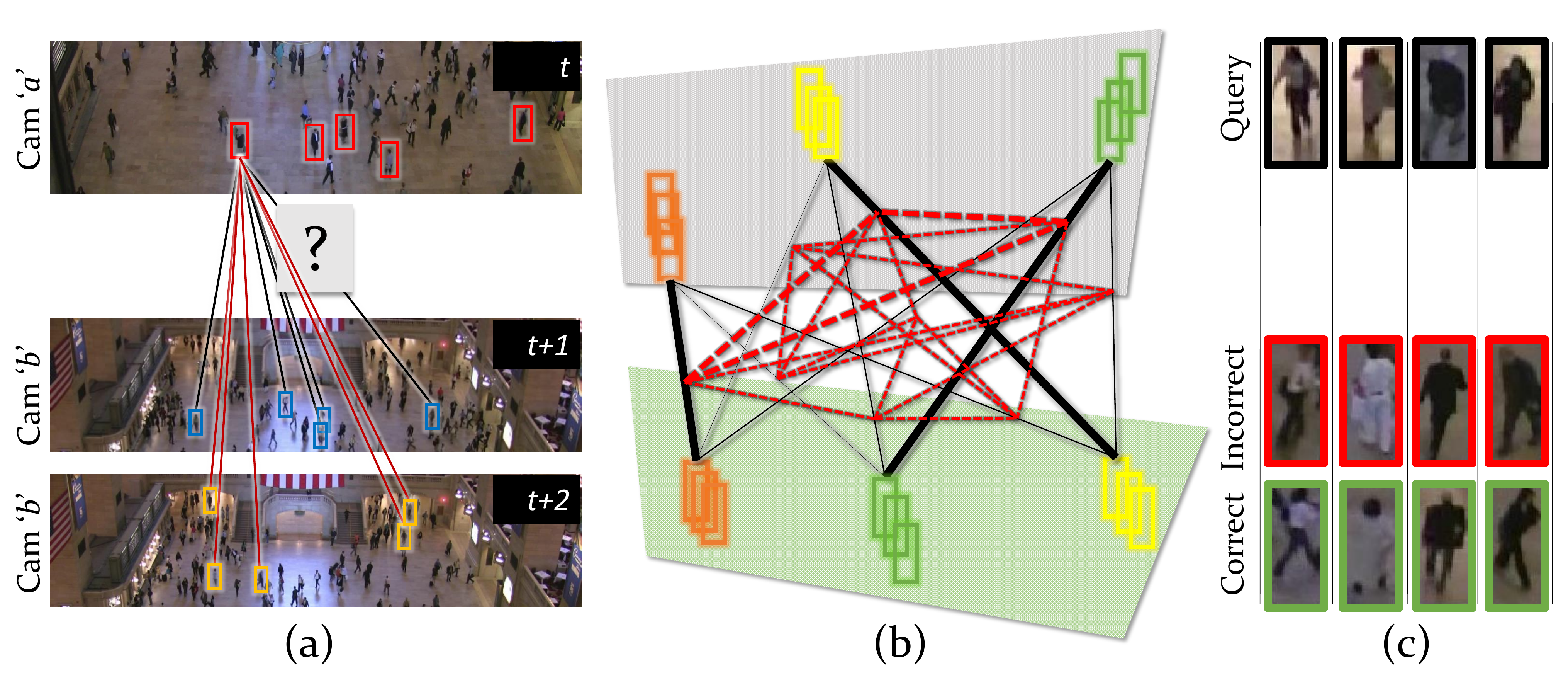}
\caption{(a) Our goal is to re-identify people leaving camera $a$ at time $t$ (top row) and when they appear in camera $b$ at some time $t+1, t+2,...$ in the future. The invisible region between the cameras is not closed, which means people can leave one camera and never appear in the other camera, or people can enter the cameras from outside without appearing in both cameras. (b) We construct a graph \textit{between individuals} in the two cameras, as shown with black lines. Some of the constraints are linear in nature (appearance, speed, destination, transition time) while others are quadratic (spatial and social grouping, collision avoidance). The quadratic constraints are shown in red and capture relationships \textit{between matches}. In (c), the people in black boxes are from camera $a$, while the other two rows shows people with similar appearance from camera $b$. The red boxes indicate the best matches (using appearance only) that are incorrect, and green boxes show the correct matches obtained by our approach with the help of other personal, social and environmental constraints. This also signifies that crowded scenes make human re-identification across cameras significantly difficult.}
\label{fig:teaser}
\end{figure*}

Human re-identification is a fundamental and crucial problem for multi-camera surveillance systems \cite{wang2013intelligent,gong2014person}. It involves re-identifying individuals after they leave field-of-view (FOV) of one camera and appear in FOV of another camera (see Fig \ref{fig:teaser}(a)). The investigation process of the Boston Marathon bombing serves to highlight the importance of re-identification in crowded scenes. Authorities had to sift through a mountain of footage from government surveillance cameras, private security cameras and imagery shot by bystanders on smart phones \cite{cnn_boston_2013}. Therefore, automatic re-identification in dense crowds will allow successful monitoring and analysis of crowded events.

Since re-identification involves associating object hypotheses, it is possible to draw some parallels to tracking as well. For non-overlapping cameras, it can be viewed as long-term occlusion handling, albeit relatively more difficult because, a) objects disappear for much longer durations of time and the observed appearances change significantly. The appearance changes due to different scene illuminations and cameras properties, and because of the objects undergoing transformations in apparent sizes and poses as well. b) Simple motion models, such as the constant velocity, also break down for longer durations of time. For instance, a person walking in a particular direction may appear as walking in a very different direction in another camera with non-overlapping FOV. Also, c) the number of possible hypotheses increases many folds since the local neighborhood priors typically employed for tracking are not applicable for re-identification.

Dense crowds are the most challenging scenario for human re-identification. For large number of people, appearance alone provides a weak cue. Often, people in crowds wear similar clothes that makes re-identification even harder. This is particularly true for the Grand Central dataset which is characterized by severe visual and appearance ambiguity among targets (Fig. \ref{fig:teaser}c). Unlike standard surveillance scenarios previously tackled in literature, we address this problem for thousands of people where at any $30$ second interval, hundreds of people concurrently enter a single camera's FOV. To handle the broad field-of-view, we divide the scene into multiple gates and learn transitions between them.

Traditionally, re-identification has been primarily concerned with matching static snapshots of people from multiple cameras. Although there have been a few works that modeled social effects for re-identification such as grouping behavior \cite{zheng2009associating,cai2010matching,bialkowski2013person}, 
they mostly deal with static images. In this paper, we study the use of time and video information for this task, and propose to consider the dynamic spatio-temporal context of individuals and the environment to improve the performance of human re-identification. The primary contribution of our work is to explicitly address the influence of personal goals, neighboring people and environment on human re-identification through high-order relationships. We complement appearance, typically employed for re-identification, with multiple personal, social and environmental (PSE) constraints, many of which are applicable without knowledge of camera topology. The PSE constraints include \textit{preferred speed and destination}, as well as \textit{social grouping} and \textit{collision avoidance}. The environmental constraints are modeled by learning the repetitive patterns that occur in surveillance networks, as individuals exiting camera from a particular location (gate) are likely to enter another camera from another specific location. The travel times between the gates are estimated as well. These are employed both as soft (\textit{spatial grouping}) and hard constraints (\textit{travel times and destination}). The PSE constraints that are linear in nature, i.e. occur between objects, are shown with black lines in Fig. \ref{fig:teaser}(b), while quadratic ones occur between matching hypotheses, i.e., pairs of objects, are shown with red lines in Fig. \ref{fig:teaser}(b). Thus, if there are $N_a$ and $N_b$ number of people in two cameras, then the total number of possible matching hypotheses is $N_aN_b$, and there are $(N_aN_b)^2$ possible quadratic hypotheses. The time limits naturally reduce some of the hypotheses, nonetheless for large number of people these constraints and costs can be overwhelming. Therefore, we propose to iteratively prune possible re-identification hypotheses in an EM-like approach, where travel times and destination probability distributions are refined using high scoring hypotheses, and hypotheses improved using updated transition information between the gates in different cameras.

We evaluate the re-identification performance using the Cumulative Matching Curve (CMC) \cite{loy2013person,zhao2013unsupervised} which quantify the rankings for each query person and average them over all queries. In addition to producing rankings for different queries, we also generate the more useful $1-1$ correspondences for individuals across cameras through joint optimization over all individuals using the proposed linear and quadratic PSE constraints. In \cite{modiri2016human}, we employed a stochastic local search algorithm to optimize the objective function simultaneously for all people. In this extension to \cite{modiri2016human}, we optimize the objective function using the efficient Frank-Wolfe algorithm \cite{frank1956algorithm,lacoste2015global} on a convex approximation of the original function with linear relaxation on the binary variables during the computation of conditional gradient, and show that the algorithm can solve the re-identification problem for hundreds of people and give an order of magnitude speed up with minor loss in performance over the stochastic local search method.

To the best of our knowledge, this is the first paper to address human re-identification using personal, social and environmental constraints in crowded scenes, and perform joint optimization to re-identify all the subjects in non-overlapping cameras. The evaluation is performed on three datasets, PRID~\cite{hirzer2011person}, DukeMTMC~\cite{ristani2016performance} and the challenging Grand~Central~dataset~\cite{yi2015understanding} which depicts dense crowds\footnote{Data and ground truth available at: \url{http://crcv.ucf.edu/projects/Crowd-Reidentification}}. Compared to our ECCV paper \cite{modiri2016human}, we make several improvements and extensions: 1) We automatically learn multi-modal distributions on transition times between gates in non-overlapping cameras, and use them to improve re-identification. 2) We train discriminative appearance model per-person using positive and negative samples mined from both cameras, and show that this boosts performance. Furthermore, 3) we tailor Frank-Wolfe algorithm to solve the re-identification problem giving an order of magnitude speed up with minor loss in performance over stochastic search algorithm. Finally, 4) we add results on DukeMTMC~\cite{ristani2016performance} dataset as well, and perform a detailed analysis of contribution of different PSE constraints to overall performance of our approach.

The rest of the paper is organized as follows. We discuss related work in Sec.~\ref{secRelatedWork}, and present the proposed personal, social and environmental (PSE) constraints in Sec.~\ref{secPSEConstraints}. The Stochastic Local Search and the computationally efficient Frank-Wolfe approaches for joint optimization over all subjects in the cameras are presented in Sec.~\ref{sec:optimization}. The results of our experiments are reported in Sec.~\ref{secExperiments}, and we conclude with some directions for future research in Sec.~\ref{secConclusion}.


\section{Related Work}\label{secRelatedWork}
Our approach is at the crossroads of human re-identification in videos, dense crowd analysis and social force models. Next, we provide a brief literature review of each of these areas.

\smallskip
\noindent\textbf{Person Re-identification} is an active area of research in computer vision, with some of the recent works including \cite{li2014deepreid,liao2015person,paisitkriangkrai2015learning,chen2015similarity,zheng2015scalable,zhang2015beyond,zhang2015group,ahmed2015improved,lisanti2015person}
applicable to static images. In videos, several methods have been developed for object handover across cameras \cite{wang2013intelligent,javed2008modeling,song2008robust,chakraborty2015network,das2014consistent}. Most of them focus on non-crowd surveillance scenarios with emphasis on modeling color distortion and learning brightness transfer functions that relate different cameras \cite{porikli2003inter,prosser2008multi,javed2005appearance,gilbert2006tracking}, whereas others relate objects by developing illumination-tolerant representations \cite{madden2007tracking} or comparing possible matches to a reference set \cite{chen2015multitarget}.
Similarly, Kuo \etal \cite{kuo2010inter} used Multiple Instance Learning to combine complementary appearance descriptors.



The spatio-temporal relationships across cameras \cite{makris2004bridging,stauffer2005learning,tieu2005inference,ardeshir2016ego2top,ardeshir2016egocentric} or prior knowledge about topology has also been used for human re-identification. Chen \etal \cite{chen2008adaptive} make use of prior knowledge about camera topology to adaptively learn appearance and spatio-temporal relationships between cameras, while Mazzon \etal \cite{mazzon2012person} use prior knowledge about relative locations of cameras to limit potential paths people can follow. Javed \etal \cite{javed2008modeling} presented a two-phase approach where transition times and exit/entrance relationships are learned first, which are later used to improve object correspondences. Fleuret \cite{fleuret2008multicamera} predicted occlusions with a generative model and a probabilistic occupancy map. Dick and Brooks \cite{dick2005stochastic} used a stochastic transition matrix to model patterns of motion within and across cameras. These methods have been evaluated on non-crowded scenarios, where observations are sparse and appearance is distinctive. In crowded scenes, hundreds of people enter a camera simultaneously within a small window of few seconds, which makes learning transition times during an unsupervised training period virtually impossible. Furthermore, since it is not always possible to obtain camera topology information, our approach is applicable whether or not the camera topology is available.

\smallskip
\noindent\textbf{Dense Crowds} studies \cite{ali2008floor,zhou2015learning} have shown that walking behavior of individuals in crowds is influenced by several constraints such as entrances, exits, boundaries, obstacles; as well as preferred speed and destination, along with interactions with other pedestrians whether moving \cite{mehran2009abnormal,ge2012vision} or stationary \cite{yi2015understanding}. Wu \etal \cite{wu2011efficient} proposed a two-stage network-flow framework for linking tracks interrupted by occlusions. Alahi~\etal~\cite{alahi2014socially} identify origin-destination (OD) pairs using trajectory data of commuters which is similar to grouping. In contrast, we employ several PSE constraints besides social grouping.

\smallskip
\noindent\textbf{Social Force Models} have been used for improving tracking performance \cite{leal2011everybody,pellegrini2009you,yamaguchi2011you}. Pellegrini \etal \cite{pellegrini2009you} were the first to use social force models for tracking. They modeled collision avoidance, desired speed and destination and showed its application for tracking. Yamaguchi \etal \cite{yamaguchi2011you} proposed a similar approach using a more sophisticated model that tries to predict destinations and groups
based on features and classifiers trained on annotated sequences. Both methods use agent-based models and predict future locations using techniques similar to crowd simulations. They are not applicable to re-identification, as our goal is not to predict but to associate hypotheses. Therefore, we use social and contextual constraints for re-identification in an offline manner. Furthermore, both these methods require observations to be in metric coordinates, which for many real scenarios might be impractical.

For re-identification in static images, group context was used by Zheng \etal \cite{zheng2009associating,gong2014person}, who proposed ratio-occurrence descriptors to capture groups. Cai \etal \cite{cai2010matching} use covariance descriptor to match groups of people, as it is invariant to illumination changes and rotations to a certain degree. For re-identifying players in group sports, Bialkowski \etal \cite{bialkowski2013person} aid appearance with group context where each person is assigned a role or position within the group structure of a team.
In videos, Qin \etal \cite{qin2013social} use grouping in non-crowded scenes to perform hand over of objects across cameras. They optimize track assignment and group detection in an alternative fashion. On the other hand, we refrain from optimizing over group detection, and use multiple PSE constraints (speed, destination, social grouping etc.) for hand over. We additionally use group context in space, i.e., objects that take the same amount of time between two gates are assigned a cost similar to grouping, when in reality they may not be traveling together in time. Mazzon and Cavallaro \cite{mazzon2013multi} presented a modified social force multi-camera tracker where individuals are attracted towards their goals, and repulsed by walls and barriers. They require a surveillance site model beforehand and do not use appearance. In contrast, our formulation avoids such assumptions and restrictions.

In summary, our approach does not require any prior knowledge about the scene nor any training phase to learn patterns of motion. Ours is the first work to incorporate multiple personal, social and environmental constraints simultaneously for the task of human re-identification in crowd videos.

\section{The Personal, Social and Environmental (PSE) Constraints}
\label{secPSEConstraints}

\begin{figure*}[t]
\centering
\includegraphics[width=1\textwidth]{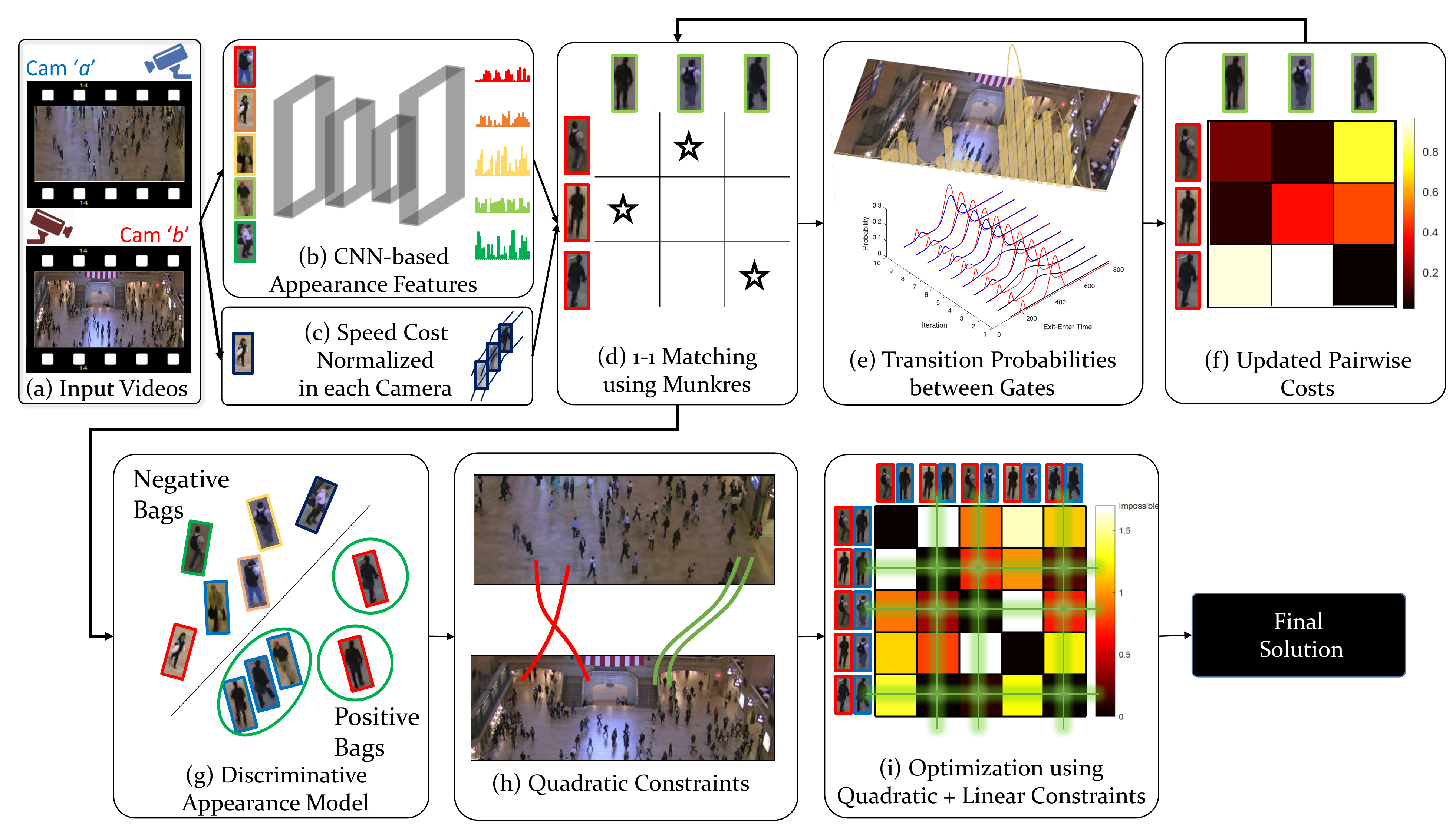}
\caption{This figure shows the pipeline of the proposed approach. (a) The input is videos from the two cameras $a$ and $b$, and human detections and tracks within those cameras. (b) Next, we extract CNN features within detection bounding boxes in the two cameras, and (c) compute similarity in speed between all possible pairs of individuals across the two cameras. (d) Both appearance and speed are used to match people, and provide initial set of re-identification hypotheses. (e) Since the transition time and destination between gates in two cameras are not known in advance, we learn them automatically using only high-confidence re-identification matches. (f) With the updated transition information between gates, the pairwise costs between individuals across cameras is re-computed using appearance, speed and consistency of re-identification hypotheses with the learned transition data.  (d-f) This process is repeated for several iterations. (g) Next, discriminative appearance models are trained per-person using positive and negative samples from both cameras, and (h) quadratic social constraints are computed between pairs of possible re-identification hypotheses. (i) Finally, we compute the $1-1$ correspondences using Stochastic Local Search or Frank-Wolfe algorithm, which yield the final solution.}
\label{fig:figPipeline}
\end{figure*}

In this section, we present our approach to re-identify people across non-overlapping cameras. We employ personal (\emph{appearance}, \emph{preferred speed}), social (\emph{spatial} and \emph{social grouping}), as well as environmental (\emph{destination}, \emph{travel time}) constraints, that are designed to be applicable even when the knowledge about camera topology is unavailable. However, when topology information is provided, additional PSE constraints (\emph{collision avoidance}, or \emph{speed in the invisible region}) become computable and are then used in our formulation (Sec. \ref{subsec:topology}). Since environmental constraints which capture transition probability distributions between different camera regions are not known a priori, we therefore solve re-identification and estimation of transition probability distributions in an alternative fashion. This is explained in Fig. \ref{fig:figPipeline} which describes the overall pipeline of our approach.


Let $O_{i_a}$ represent an observation of an object $i$ in camera $a$. Its trajectory (track) is given by a set of points $[\v p_{i_a}(t^{\eta}_{i_a}), \dots, \v p_{i_a}(t^{\chi}_{i_a})]$, where $t^{\eta}_{i_a}$ and $t^{\chi}_{i_a}$ represent the time it entered and exited the camera $a$, respectively. Given another observation of an object $j$ in camera $b$, $O_{j_b}$, a possible match between the two is denoted by ${M_{i_a}^{j_b}} = {\langle O_{i_a}, O_{j_b} \rangle}$. To simplify notation, we drop the symbol for time $t$ and use it only when necessary, thus, $\v p^{\chi}_{i_a} \equiv \v p_{i_a}(t^{\chi}_{i_a})$ and $\v p^{\eta}_{j_b} \equiv \v p_{j_b}(t^{\eta}_{j_b})$.

The entrances and exits in each camera are divided into multiple gates. These gates are virtual in nature, and correspond to different regions in the fields-of-view of cameras.  For the case of two cameras $a$ and $b$, the gates are given by $\v G_{1_a},\ldots,\v G_{U_a}$ and $\v G_{1_b},\ldots,\v G_{U_b}$, where $U_a$ and $U_b$ are the total number of gates in both cameras, respectively. Furthermore, we define a function $g(\v p(t))$, which returns the nearest gate when given a point in the camera. For instance, for a person $i_a$, $g(\v p^{\chi}_{i_a})$ returns the gate from which the person $i$ exited camera $a$, by computing the distance of $\v p^{\chi}_{i_a}$ to each gate. Mathematically, this is given by:
\begin{equation}
g(\v p^{\chi}_{i_a}) = \argmin_{\v G_{u_a}}\|\v G_{u_a} - \v p^{\chi}_{i_a}\|^2, \;\; \forall u_a=1,\ldots,U_a.
\end{equation}

Next, we describe the costs for different PSE linear, $\phi(.)$, and quadratic, $\varphi(.)$, constraints employed in our framework for re-identification. Since all costs have their respective ranges, we use a logistic function, $\hat{\phi}(.) = \alpha(1 + \exp(-\beta \phi(.))^{-1}$, to balance them. Most of the constraints do not require knowledge about camera topology, and are described next.

\subsection{Personal Constraints}\label{subsec:personal}

The personal constraints, which are linear in nature, capture the individual characteristics in the form of appearance and motion of each person in the different cameras.


\smallskip
\noindent\textbf{Appearance:} To compute appearance similarity between observations $O_{i_a}$ and $O_{j_b}$, we use features from Convolutional Neural Networks \cite{simonyan2014very}. In particular, we extract features from Relu6 and Fc7 layers, followed by homogenous kernel mapping \cite{vedaldi2012efficient} and linear kernel as the the similarity metric. However, computing appearance similarity using single snapshot (bounding box) is suboptimal, as overtly visible background in the detection bounding box, or occlusions and noise can cause a drop in performance. To handle this, we sample five snapshots per track, i.e., multiple detections along the track, and then take the median of the appearance similarity between all possible 5x5 detection pairs as overall similarity between two observations $O_{i_a}$ and  $O_{j_b}$. Since median is less sensitive to outliers, we found it to outperform minimum and maximum functions. This yields appearance similarity, $\phi_{\textrm{app}}(O_{i_a}, O_{j_b})$ between objects $O_{i_a}$ and $O_{j_b}$.

However, an important observation regarding videos is that they provide spatio-temporal information about the observed individuals which can be used to learn strong discriminative appearance models. This stems from the fact that it is always possible to find a set of people which the individual under consideration can \textit{never} match within the same camera, and more importantly, in the other cameras. For instance, consider $O_{i_a}$ in camera $a$ for which we intend to learn a discriminative model. Let its appearance descriptors be given by $[\v x_{i_a}(t^{\eta}_{i_a}), \dots, \v x_{i_a}(t^{\chi}_{i_a})]$ corresponding to the points $[\v p_{i_a}(t^{\eta}_{i_a}), \dots, \v p_{i_a}(t^{\chi}_{i_a})]$ from its track. To simplify notation, we drop time and represent multiple snapshots of an individual available in its track with $\v x_{i_a}$. The discriminative model requires both positive and instances samples, and with multiple cameras, we get four cases:

\begin{itemize}
  \item{$S^{+}_a$: This includes all the samples from the track of individual $O_{i_a}$ in camera $a$, i.e., $\v x_{i_a}$ with label $y_{i_a}=1$.}
  \item{$S^{-}_a$: These include all the samples from tracks of other individuals in camera $a$, i.e., $\{\v x_{i'_a}, \forall i'_a | i'_a \neq i_a$\} with corresponding labels $y_{i'_a}=-1$.}
  \item{$S^{-}_b$: These include individuals in camera $b$, $\{O_{j_b}\}$ that can never match to $O_{i_a}$. For instance, individuals from the past, or way into the future, or individuals that are extremely poor matches using other easy-to-compute costs. When the number of people is large, only hard negatives are used. Each instance in this set has label $y_{j_b}=-1$.}
  \item{$S^{+}_b$: Since we do not know the true positive or re-identification match in camera $b$, we relax this constraint by ensuring a single positive label over multiple possible matches, i.e., individuals $O_{j_b}$ which have low appearance and speed costs and lie within the expected time frame. Only one of the instances in this set can have a positive label, therefore $\sum_{j_b \in S^{+}_b} (y_{j_b}+1)/2 = 1$.}
\end{itemize}

Thus, the objective function for the soft-margin classifier becomes:
\begin{align}
& \min_{\v w, v, \xi} \frac{1}{2}\|\v w\|^2 + C\sum_{i}\xi_{i} & \label{eqMISVMObj} \\
\textmd{s.t.}\;\;\;
& y_{i}(\langle \v w, \v x_{i} \rangle + v) \geq 1 - \xi_{i}, & \forall i \in S^{+}_a, S^{-}_a, S^{-}_b, \label{eqMISVMConst1}\\
& \max(\langle \v w, \v x_{i} \rangle + v) \geq 1 - \xi_{i}, & \forall i \in S^{+}_b, \label{eqMISVMConst2}\\
& \xi_{i} \geq 0,\;\; y_i \in \{-1,1\}, \label{eqMISVMConst3}
\end{align}
where $\v w, v$ are SVM weight vector and bias respectively, and $\xi$ represents the slack variables. Note that in Eq. \ref{eqMISVMConst2}, $y_i = +1$ and is therefore omitted. This forms a special case of Multiple Instance Learning \cite{andrews2002support}, solved by imputing the labels of $S^{+}_b$ and solving the SVM objective, alternatively. In essence, the above formulation takes advantage of video information using semi-labeled data, and also handles the issue of domain adaptation by forcing the classifier to perform well on hard negatives from the other camera. The cost between two objects using discriminative appearance model is given by:
\begin{equation}
\phi_{\textrm{disc}}(O_{i_a}, O_{j_b}) = \langle \v w_{i_a}, \v x_{j_b} \rangle + v_{i_a}.
\end{equation}

\smallskip
\noindent\textbf{Preferred Speed:} The walking speed of humans has been estimated to be around $1.3$ m/s \cite{robin2009specification}. Since, we do not assume the availability of metric rectification information, we cannot use this fact directly in our formulation. However, a consequence of this observation is that we can assume the walking speed of individuals, \textit{on average}, in different cameras is constant. We assume a Normal distribution, $\mathcal N(.)$, on observed speeds in each camera. The variation in walking speeds of different individuals is captured by the variance of the Normal distribution. Let $\mathcal N(\mu_a, \sigma_a)$ and $\mathcal N(\mu_b, \sigma_b)$ denote the distribution modeled in the two cameras. Since a particular person is being assumed to walk with the same speed in different cameras, the cost for preferred speed using the exit speed of person $i_a$, $\dot{\v p}^{\chi}_{i_a}$, and the entrance speed of person $j_b$, $\dot{\v p}^{\eta}_{j_b}$ is given by:
\begin{gather}\label{eqSpeedCost}
\dot{\v p}^{\chi}_{i_a} = \sigma_a^{-1} ( \|\v p^{\chi}_{i_a} - \v p^{\chi - 1}_{i_a} \| - \mu_a),\;\;\\
\dot{\v p}^{\eta}_{j_b} = \sigma_b^{-1} ( \|\v p^{\eta+1}_{j_b} - \v p^{\eta}_{j_b} \| - \mu_b),\\
\phi_{\textrm{spd}}(O_{i_a}, O_{j_b}) = |\dot{\v p}^{\chi}_{i_a} - \dot{\v p}^{\eta}_{j_b}|.
\end{gather}

%


\subsection{Environmental Constraints}\label{subsec:environmental}

Next, we describe the environmental constraints, which are linear in nature and predict the most probable paths and travel times between gates across cameras.

\smallskip
\noindent\textbf{Destination and Travel Time:} For re-identification in multiple cameras, the knowledge about probable destination gives a prior for an individual's location in another camera. Furthermore, since people disappear between cameras, the consistency in time required to travel between a particular set of gates in two different cameras for different individuals serves as an implication to their correctness. We capture these observations by modeling the transition probability distributions between gates in different cameras, as well the time required to travel between them.

\begin{figure*}[t]
\centering
\includegraphics[width=1\textwidth]{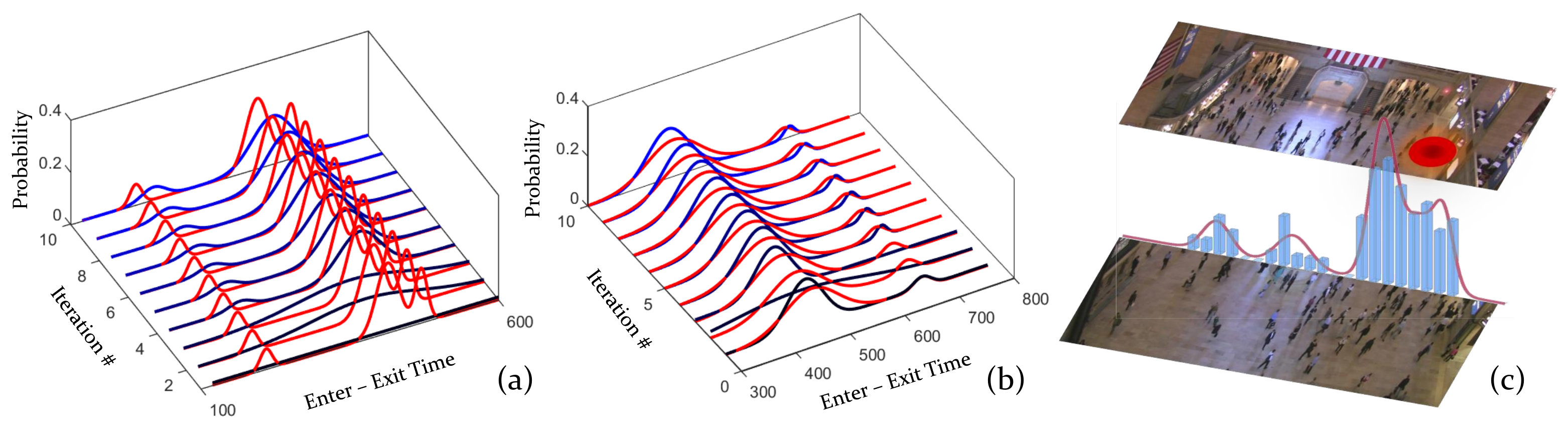}
\caption{This figure shows the intermediate results while computing transition probability distributions between pairs of virtual gates in Grand Central dataset. (a,b) show two examples of estimation of travel times for pairs of gates approximated automatically using Eq. \ref{eqTravelTimesCost}. Here, red curves show the ground truth distribution of travel time, whereas black and blue curves show automatically learned travel time distribution across different iterations. (c) shows the destination probability distribution, estimated using Eq. \ref{eqDestinationCost}, for people exiting a particular gate in one camera (shown with red circle) with respect to entrance gates in the other camera.}
\label{fig:figTransition}
\end{figure*}

Assuming we have a set of putative matches $\{M_{i_a}^{j_b}\}$ (Fig. \ref{fig:figPipeline} (d)), we estimate the probability distribution of transition between exit gate $G_{u_a}$ and entrance gate $G_{u_b}$ as:
\begin{equation}\label{eqDestinationCost}
p(G_{u_a},G_{u_b})
= \frac{| g(\v p^{\chi}_{i_a})=G_{u_a} \wedge g(\v p^{\eta}_{j_b})=G_{u_b} |} {\sum_{u'_b} | g(\v p_{i_a}^{\chi})=G_{u_a} \wedge g(\v p_{j_b}^{\eta})=G_{u'_b}| },
\end{equation}
while the travel times are modeled using Mixture-of-Gaussians \cite{figueiredo2002unsupervised} for each pair of gates. We use up to $K=5$ components, where exact number is automatically determined using data by \cite{figueiredo2002unsupervised}. Thus, the probability of travel time is given by,
\begin{equation}\label{eqTravelTimesCost}
q(\Delta t | G_{u_a},G_{u_b}) = \sum_{k=1}^{K} w_k\mathcal{N}(\mu_k, \Sigma_k).
\end{equation}

Thus, the cost for destination and travel times between gates for the match $\langle O_{i_a}, O_{j_b} \rangle$ is given by:
\begin{align}\label{eqDestTimeCost}
\hspace{3em}&\hspace{-3em}\phi_{\textrm{tr}}(O_{i_a}, O_{j_b}) = \nonumber\\
&- p\big(g(\v p_{i_a}^{\chi}), g(\v p_{j_b}^{\eta}) \big)
\cdot q(t^{\eta}_{j_b} - t^{\chi}_{i_a} | g(\v p_{i_a}^{\chi}), g(\v p_{j_b}^{\eta}).
\end{align}

Since the transition probability distributions in Eq. \ref{eqDestinationCost} and travel times in Eq. \ref{eqTravelTimesCost}
are not known in advance, we use an EM-like approach that iterates between solving $1-1$ correspondences using the linear and quadratic constraints, and estimating transition information using those correspondences (Fig. \ref{fig:figPipeline} (d-f)). Fig. \ref{fig:figTransition} shows some intermediate results when computing transition time and destination probability distributions. For travel times, we initialize with the uniform distribution, and update the travel time distribution with a momentum of 0.15 at each iteration. As can be seen with blue curves in Fig. \ref{fig:figTransition} (a,b), the estimation of travel times improves across iterations.

\subsection{Social Constraints}\label{subsec:social}

The quadratic social constraints are computed between pairs of re-identification hypotheses, i.e., between possible matches that have the same destinations and travel times.

\smallskip
\noindent\textbf{Spatial Grouping:} The distance traveled by different individuals between two points (or gates) across cameras should be similar. Since the camera topology is not available in this case, the distance can be implicitly computed as a product of velocity and time. This is a quadratic cost computed between every two possible matches, $M_{i_a}^{j_b}$ and $M_{i'_a}^{j'_b}$, given by:
\begin{align}\label{eqSpatial}
\hspace{3em}&\hspace{-3em}\varphi_{\textrm{spt}}(M_{i_a}^{j_b}, M_{i'_a}^{j'_b}) = \nonumber\\
&\exp(-|\v p^{\chi}_{i_a} - \v p^{\chi}_{i'_a}|) \cdot \exp(-|\v p^{\eta}_{j_b} - \v p^{\eta}_{j'_b}|)\nonumber\\
&\cdot | (\dot{\v p}^{\chi}_{i_a}+\dot{\v p}^{\eta}_{j_b})(t^{\eta}_{j_b} - t^{\chi}_{i_a})  - (\dot{\v p}^{\chi}_{i'_a}+\dot{\v p}^{\eta}_{j'_b})(t^{\eta}_{j'_b} - t^{\chi}_{i'_a})|.
\end{align}

Effectively, if the exit and entrance locations are nearby (the first two terms in Eq. \ref{eqSpatial}), then we compute the distance traveled by each match in the pair using the product of mean velocity and time required to travel between those locations (the third term). It is evident from Eq. \ref{eqSpatial} that the exponentiation in first two terms will allow this cost to take effect only when the entrance and exit locations are both proximal. If so, the third term will then measure the difference in distance traveled by the two possible matches (tracks), and penalize using that difference. If the distance is similar, the cost will be low suggesting both matches (tracks) should be included in the final solution. If the difference is distance is high, then at least one or both of the matches are incorrect.

\smallskip
\noindent\textbf{Social Grouping:} People tend to walk in groups. In our formulation, we reward individuals in a \textit{social group} that exit and enter together from the same locations at the same times,
\begin{align}\label{eqGroup}
\hspace{1em}&\hspace{-1em}\varphi_{\textrm{grp}}(M_{i_a}^{j_b}, M_{i'_a}^{j'_b}) =\nonumber\\
&\exp(-|\v p^{\chi}_{i_a} - \v p^{\chi}_{i'_a}|-|\v p^{\eta}_{j_b} - \v p^{\eta}_{j'_b}|-|t^{\eta}_{j_b} - t^{\eta}_{j'_b}|- |t^{\chi}_{i_a} - t^{\chi}_{i'_a}|).
\end{align}

Here, the first two terms capture the difference in exit and entrance locations, respectively, and the third and fourth terms capture the difference in exit and entrance times, respectively.

\subsection{PSE Constraints with Camera Topology}\label{subsec:topology}

The PSE constraints presented in the previous subsections are applicable when the spatial relations between the cameras are not known. However, if the inter-camera topology is available, then it can be used to infer the motion of people as they travel in the invisible or unobserved regions between the cameras. The quality of paths in the invisible region can be subject to constraints such as \textit{preferred speed} or \textit{direction of movement}, which can be quantified and introduced into the framework. Furthermore, collision avoidance is another social constraint that can only be applied when inter-camera topology is known.

Given two objects in cameras $a$ and $b$, $O_{i_a}$ and $O_{i_b}$, in the same reference of time, we predict the possible trajectory between the object hypotheses. This is obtained by fitting a spline, given by $\bm \gamma_{i_a}^{j_b}$, in both $x$ and $y$ directions using cubic interpolation between the points $\v p_{i_a}$ and $\v p_{j_b}$ parameterized with their respective time stamps.

\smallskip
\noindent\textbf{Collision Avoidance:} Let the point of closest approach between two interpolated trajectories be given by:
\begin{equation}
d(\bm \gamma_{i_a}^{j_b}, \bm \gamma_{i'_a}^{j'_b}) = \min_{\max(t_{i_a}^{\chi}, t_{i'_a}^{\chi} ),\ldots,\min(t_{j_b}^{\eta}, t_{j'_b}^{\eta})} \|\bm \gamma_{i_a}^{j_b}(t) - \bm \gamma_{i'_a}^{j'_b}(t)\|,
\end{equation}
we quantify the collision avoidance as a quadratic cost between pairs of possible matches:
\begin{multline}\label{eqAccCostTopology}
\phi_{\textrm{invColl}}(M_{i_a}^{j_b}, M_{i'_a}^{j'_b}) = \\\big( 1 - \varphi_{\textrm{grp}}(M_{i_a}^{j_b}, M_{i'_a}^{j'_b}) \big)
.\exp \big(- d(\bm \gamma_{i_a}^{j_b}, \bm \gamma_{i'_a}^{j'_b}) \big).
\end{multline}

Since people avoid collisions with others and change their paths, this is only applicable to trajectories of people who are not traveling in a group (first term in Eq. \ref{eqAccCostTopology}), i.e., the cost will be high if two people not walking in a group come very close to each other when traveling through the invisible region between the cameras.

\smallskip
\noindent\textbf{Speed in Invisible Region:} The second constraint we compute is an improved version of the \textit{preferred speed} - a linear constraint which now also takes into account the direction is addition to speed of the person in the invisible region. If the velocity of a person within visible region in cameras and while traveling through the invisible region is similar, this cost would be low. However, for an incorrect match, the difference between speed in visible and invisible regions will be high. Let $\dot{\bm \gamma}$ denote the velocity at respective points in the path, both in the visible and invisible regions. Then, the difference of maximum and minimum speeds in the entire trajectory quantifies the quality of a match, given by,
\begin{equation}\label{eqSpeedCostTopology}
\phi_{\textrm{invSpd}}(O_{i_a}, O_{j_b}) = | \max_{t_{i_a}^{\eta} \ldots t_{j_b}^{\chi}}{\dot{\bm \gamma}_{i_a}^{j_b}(t)}
- \min_{t_{i_a}^{\eta} \ldots t_{j_b}^{\chi}}{\dot{\bm \gamma}_{i_a}^{j_b}(t)}|.
\end{equation}


When the inter-camera topology is available, these constraints are added to the Eq. \ref{eqL} and the method described in the Sec. \ref{sec:optimization} is used to re-identify people across cameras.


\section{Optimization with PSE Constraints}\label{sec:optimization}
In this section, we present the optimization techniques which use the aforementioned constraints. Let $z_{i_a}^{j_b}$ be the variable corresponding to a possible match $M_{i_a}^{j_b}$. Our goal is to optimize the loss function over all possible matches and pairs of matches, which is the weighted sum of linear and quadratic terms. When knowledge about topology is not available, the loss function is given by:


\begin{align}\label{eqL}
& \nonumber \mathcal{L}(\v z) = \sum_{\substack{{i_a, j_b}\\ {i'_a, j'_b}}} z_{i_a}^{j_b} z_{i'_a}^{j'_b} \underbrace{ \Big( \hat{\varphi}_{\textrm{spt}}(M_{i_a}^{j_b}, M_{i'_a}^{j'_b}) + \hat{\varphi}_{\textrm{grp}}(M_{i_a}^{j_b}, M_{i'_a}^{j'_b}) \Big)}_{\v Q (\text{Quadratic Terms})} + \\
& \sum_{i_a,j_b} z_{i_a}^{j_b} \underbrace{\big( \hat{\phi}_{\textrm{app}}(M_{i_a}^{j_b}) + \hat{\phi}_{\textrm{disc}}(M_{i_a}^{j_b}) + \hat{\phi}_{\textrm{spd}}(M_{i_a}^{j_b}) + \hat{\phi}_{\textrm{tr}}(M_{i_a}^{j_b})\big)}_{\v L (\text{Linear Terms})},
\end{align}
subject to the following conditions:
\begin{equation}\label{eqConstraints}
\sum_{i_a}z_{i_a}^{j_b} \leq 1, \forall j_b,
\sum_{j_b}z_{i_a}^{j_b} \leq 1, \forall i_a,
z_{i_a}^{j_b} \in \{0,1\}.
\end{equation}




The problem in Eq. \ref{eqL} is non-convex due to the nature of PSE constraints, and due to binary and combinatorial nature of variables in Eq. \ref{eqConstraints}, it is NP-hard. The first solution we present solves the non-convex optimization in its original form (presented in our ECCV paper \cite{modiri2016human}) through Stochastic Local Search, and a new approach which uses convex approximation on the loss function and linear relaxation on the binary constraints when computing conditional gradient, and is solved using the Frank-Wolfe algorithm \cite{frank1956algorithm,lacoste2015global}.

\subsection{Stochastic Local Search Optimization} \label{subsec:StochasticSearch}
\begin{algorithm*}[t]
\caption{: Algorithm to find $1-1$ correspondence between persons observed in different cameras using both linear and quadratic constraints.}
\label{alg:optimization}

\textbf{Input}: $O_{i_a}, O_{j_b} \;\;\; \forall i_a, j_b$, $R$ (\# steps)  \\
\textbf{Output}: $\mathcal{L}^*, \v z^*; \;\; \;\; 0 \leq |t^{\eta}_{j_b} - t^{\chi}_{i_a}| \leq \tau, \forall z_{i_a}^{j_b}$

\rule{1\linewidth}{0.02cm}

\begin{algorithmic}[1]
\Procedure{Re-Identify}{$ $}
	\State Initialize $[\mathcal{L}^*, \v z^*]$ for Linear Constraints with $\textproc{Munkres}$  \cite{munkres1957algorithms} \Comment{Initial solution}

    \While {$\mathcal{L}^*$ improves}
        \For {$r=0$ to $R$} 
     	    \State $[\mathcal{L}^-, \v z^-] = \textproc{RemoveMat}(\mathcal{L}^*, \v z^*, r)$ \Comment{Probabilistically remove $r$ matches}
            \State $\mathcal{L}' = \mathcal{L}^-, \v z' = \v z^-$ \Comment{Consider it the new solution}
            \For {$s=r+1$ to $1$} 
                \State $[\mathcal{L}^+, \v z^+] = \textproc{AddMat}(\mathcal{L}', \v z', s)$ \Comment{Add $s$ new matches to the solution}
                \If {$\mathcal{L}' > \mathcal{L}^+$} \Comment{Is the solution after adding new matches better?}
                    \State $\mathcal{L}' = \mathcal{L}^+, \v z' = \v z^+$ \Comment{If so, update it as the new solution}
                \EndIf
            \EndFor
            \If {$\mathcal{L}^* > \mathcal{L}'$} \Comment{Is the new solution better the best solution so far?}
                \State $\mathcal{L}^* = \mathcal{L}', \v z^* = \v z'$ \Comment{If so, update it as the best solution}
            \EndIf
        \EndFor
     \EndWhile
\EndProcedure
\end{algorithmic}
\end{algorithm*}

Algorithm \ref{alg:optimization} presents an approach which optimizes Eq. \ref{eqL} subject to the conditions in Eq. \ref{eqConstraints} through Stochastic Local Search \cite{spall2005introduction}. The solution is initialized using linear terms with Munkres \cite{munkres1957algorithms}. This is followed by stochastic removal and addition of matches into the current solution. However, the solution is updated whenever there is a decrease in the loss function in Eq. \ref{eqL}, as can be seen from Line $13$. Once the change in loss is negligible or a maximum pre-defined number of iterations is reached, the algorithm stops and returns the best solution obtained. In Alg. \ref{alg:optimization}, the sub-procedure $\textproc{RemoveMat}(\mathcal{L}, \v z, r)$ removes $r$ hypotheses from the solution as well as their respective linear and quadratic costs by assigning probabilities (using respective costs) for each node in the current solution $\v z$. In contrast, the sub-procedure $\textproc{AddMat}(\mathcal{L}, \v z,s)$ adds new hypotheses to the solution using the following steps:

\begin{itemize}
  \itemsep0em
  \item Populate a list of matches for which $z_{i_a}^{j_b}$ can be $1$ such that Eq. \ref{eqConstraints} is satisfied.
  \item Generate combinations of cardinality $s$ using the list.
  \item Remove combinations which dissatisfy Eq. \ref{eqConstraints}.
  \item Compute new $\mathcal{L}$ in Eq. \ref{eqL} for each combination. This is efficiently done by adding $|\v z|*s$ quadratic values and $s$ linear values.
  \item Pick the combination with lowest loss $\mathcal{L}$, add it to $\v z$ and return.
\end{itemize}

\begin{figure}[t]
\centering
\includegraphics[width=0.5\textwidth]{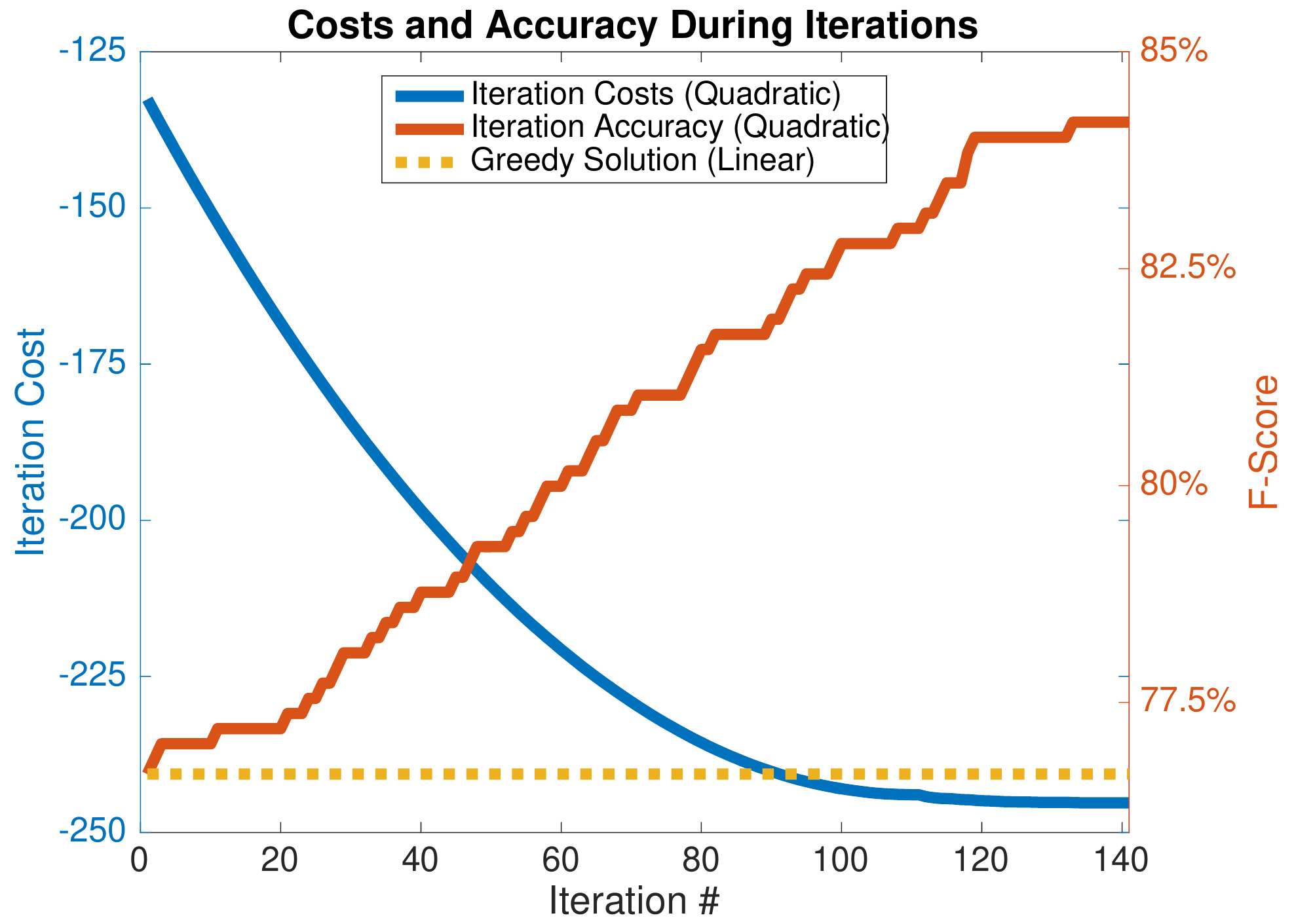}
\caption{The graph shows the performance of Algorithm \ref{alg:optimization} using both linear and quadratic constraints, compared against Hungarian Algorithm \cite{munkres1957algorithms} using only the linear costs shown with orange dotted line. The loss function in Eq. \ref{eqL} is shown in blue, whereas the accuracy is shown in red. Quadratic PSE constraints in conjunction with Alg. \ref{alg:optimization} yield an improvement of  \textbf{$\sim8\%$} over linear constraints.}
\label{fig:intermediateOptimization}
\end{figure}

Fig. \ref{fig:intermediateOptimization} shows the intermediate results for this optimization approach using Alg. \ref{alg:optimization}. The $x$-axis is the step number, whereas the left $y$-axis shows the value of loss function in Eq. \ref{eqL} (blue curve), and the right $y$-axis shows the F-Score in terms of correct matches (orange curve). We also show results of Hungarian Algorithm (Munkres) \cite{munkres1957algorithms} in dotted orange line using linear constraints, which include appearance and speed similarity. These curves show that Alg. \ref{alg:optimization} simultaneously improves the loss function in Eq. \ref{eqL} and the accuracy of the matches as the number of steps increases.

\subsection{Frank-Wolfe Optimization}\label{subsec:FW}

The Stochastic Local Search algorithm presented in the previous subsection optimizes over a non-convex quadratic function with binary variables. Rewriting the loss function in matrix form in Eq. \ref{eqL}, we get
\begin{equation}\label{eqL2}
\mathcal{L}(\v z) = \v z^T \v Q \v z + \v L \v z,
\end{equation}
subject to the linear and binary constraints in Eq. \ref{eqConstraints}. Due to the linear nature of the constraints, the convex hull or polytope $\mathcal{D}$ from Eq. \ref{eqConstraints} is convex. Furthermore, the quadratic function in Eq. \ref{eqL2} can be made convex by taking the normalized Laplacian of $\v Q$, $\hat{\v Q} = \v I - \v D^{\frac{1}{2}} \v Q \v D^{\frac{1}{2}}$, where the diagonal matrix $\v D$ contains the row sums of $\v Q$ and $\v I$ is the identity matrix. This allows the use of Frank-Wolfe algorithm (also known as Conditional Gradient Method) which approximates Eq. \ref{eqL2} with linear subproblems and iteratively minimizes the following objective:
\begin{equation}\label{eqFWhComplete}
\v h_{k+1} = \underset{\v h \in \mathcal{D}}{\textrm{arg\;min}}\;\; \mathcal{L}(\v z_{k}) + \nabla\mathcal{L}(\v z_{k})^{T}(\v h - \v z_{k}).
\end{equation}

Since the minimization in Eq. \ref{eqFWhComplete} does not depend on $\v z_k$, we get the following equivalent optimization problem:
\begin{equation}\label{eqFWh}
\v h_{k+1} = \underset{\v h \in \mathcal{D}}{\textrm{arg\;min}}\;\; \langle \v h, \nabla\mathcal{L}(\v z_{k}) \rangle,
\end{equation}
and the solution is updated as a weighted average of previous and new solution:
\begin{equation}\label{eqFWUpdate}
\v z_{k+1} = (1 - \lambda_{k+1}) \v z_k + \lambda_{k+1} \v h_{k+1},
\end{equation}
where
\begin{equation}\label{eqFWLambda}
\lambda_{k+1} = \underset{\lambda \in [0,1]}{\textrm{arg\;min}} \;\;  \mathcal{L}(\v z_k + \lambda(\v h_{k+1} - \v z_k)).
\end{equation}

Typically, $\lambda_{k+1} = 2/(2+k)$ such that the weight of new solution reduces as number of iterations increase, or it can be computed through line search. Since Eq. \ref{eqL2} is quadratic, a closed-form solution to $\lambda_{k+1}$ exists. Using the fact that $\nabla \mathcal{L}(\v z) = -\v Q \v z - \v L$, the minimum values of Eq. \ref{eqFWLambda} occurs when,
\begin{gather}
\frac{\partial}{\partial \lambda} \mathcal{L}\big(\v z_k + \lambda_{k+1}(\v h_{k+1} - \v z_k)\big) = 0,\\
\nabla \mathcal{L}\big(\v z_k + \lambda_{k+1}(\v h_{k+1} - \v z_k)\big)^T(\v h_{k+1} - \v z_k) = 0,\\
\lambda_{k+1} = \frac{\nabla \mathcal{L}(\v z_k)^T (\v h_{k+1} - \v z_k)}{(\v h_{k+1} - \v z_k)^T \v Q (\v h_{k+1} - \v z_k)}.
\end{gather}

The algorithm begins by randomly initializing a point $\v h_0$ from the solution space $\mathcal{D}$. Given the current solution $\v z_{k}$, Eq. \ref{eqFWh} finds the point where the gradient of the loss function is minimum, and the solution is updated through Eq. \ref{eqFWUpdate}. For the case of linear program, the solution lies on the boundary of the polytope. To avoid the occasional zig-zag behavior of optimizer for solutions near boundary of the polytope, an \textit{away} step is computed within the convex hull of points in $\mathcal{D}$ visited till iteration $k$, i.e. $\mathcal{S}_{k}$:
\begin{gather}
\v f_{k} \leftarrow \underset{\v h \in \mathcal{S}_{k}}{\textrm{argmax}} \langle \v h, \nabla\mathcal{L}(\v z) \rangle\label{eqFWf},
\end{gather}
where $\mathcal{S}_{k}$ contains the \textit{active corners} or previously seen integer solutions. Then, at each iteration, the step that gives the steepest descent between Eq. \ref{eqFWh} and Eq. \ref{eqFWf} is selected. Since the solution $\v z_{k+1}$ of Frank-Wolfe at each iteration does not satisfy the binary constraints in Eq. \ref{eqConstraints}, we round the solution in $\v z_{k+1}$ by transforming the vector $\v z_{k+1}$ into a matrix, where rows correspond to people in the first camera, and columns to people in second camera. The new cost matrix is solved through \textsc{Munkres} for $1-1$ matching and the solution satisfies Eq. \ref{eqConstraints}. For the next iteration of the Frank-Wolfe, we transform the binary matrix again into vector form where each element represents a re-identification hypothesis. This process of transforming the variables to binary is termed as \emph{rounding}. The algorithm is run till the duality gap is larger than a fixed threshold or a pre-defined number of iterations is reached.

Compared to Stochastic Local Search (Alg. \ref{alg:optimization}), Frank-Wolfe solves a convex approximation of the original objective function in Eq. \ref{eqL}, and relaxes the binary constraints in Eq. \ref{eqConstraints} when computing the conditional gradient. As we will see in Sec. \ref{secExperiments}, this results in a slight drop in performance with an order of magnitude gain in computation speed.

\section{Experiments}\label{secExperiments}

Since PSE constraints depend on time and motion information in the videos, many commonly evaluated datasets such as VIPeR \cite{gray2008viewpoint} and ETHZ \cite{ess2008mobile} cannot be used for computing PSE constraints. We evaluate the proposed approach on the PRID~\cite{hirzer2011person}, DukeMTMC~\cite{ristani2016performance} and the challenging Grand Central Dataset \cite{yi2015understanding}. First, we introduce the datasets and the ground truth that was generated for evaluation, followed by detailed analysis of our approach as well as contribution of different personal, social and environmental (PSE) constraints to the overall performance.

\subsection{Datasets and Experimental Setup}\label{subsection:Dataset}

\subsubsection{Grand Central}
Grand Central is a dense crowd dataset that is particularly challenging for the task of human re-identification.
The dataset contains $120,000$ frames, with a resolution of $1920 \times 1080$ pixels. Recently, Yi \etal \cite{yi2015understanding} used a portion of the dataset for detecting stationary crowd groups. They released annotations for trajectories of $12,684$ individuals for $6,000$ frames at $1.5$ fps. We rectified the perspective distortion from the camera and put bounding boxes at correct scales using the trajectories provided by \cite{yi2015understanding}. However, location of annotated points were not consistent for any single person, or across different people. Consequently, we manually adjusted the bounding boxes for $1,500$ frames at $1.5$ fps, resulting in ground truth for $17$ minutes of video data.

We divide the scene into three horizontal sections, where two of them become separate cameras and the middle section is treated as invisible or unobserved region. The locations of people in each camera are in independent coordinate systems. The choice of dividing the scene in this way is meaningful, as both cameras have different illuminations due to external lighting effects, and the size of individuals is different due to perspective effects. Furthermore, due to the wide field of view in the scene, there are multiple entrances and exits in each camera, so that a person exiting the first camera at a particular location has the choice of entering from multiple different locations. Figure \ref{fig:teaser}(c) shows real examples of individuals from the two cameras and elucidates the fact that due to the low resolution, change in brightness and scale, the incorrect nearest neighbors matches using the appearance features often rank higher than the correct ones for this dataset.

\subsubsection{DukeMTMC}
Recently, the DukeMTMC dataset was released to quantify and evaluate the performance of multi-target, multi-camera tracking systems. It is high resolution 1080p, 60fps dataset and includes surveillance footage from 8 cameras with approximately 85 minutes of videos for each camera. There are cameras with both overlapping and non-overlapping fields-of-view. The dataset is of low density with 0 to 54 people per frame. Since only the ground truth for training set has been released so far, which constitutes first 50 minutes of video for each camera, we report performance on the training set only. Cameras 2 and 5 which are disjoint, and have the most number of people (934 in total, with 311 individuals appearing in both cameras), were selected for experiments. To remain consistent with the other datasets, we perform evaluation in terms of Cumulative Matching Curves (CMC) and F-Score on 1-1 assignment.

\subsubsection{PRID}
PRID 2011 is a camera network re-identification dataset containing 385 pedestrians in camera `\textit{a}' and $749$ pedestrians in camera `\textit{b}'. The first $200$ pedestrians from each camera form the ground truth pairs while the rest appear in one camera only. The most common evaluation method on this dataset is to match people from cam `\textit{a}' to the ones in cam `\textit{b}'. We used the video sequences and the bounding boxes provided by the authors of \cite{hirzer2011person} so we can use the PSE constraints in our evaluation. Since the topology of the scene is unknown, we have used the constraints which do not need any prior knowledge about the camera locations. We evaluated on the entire one hour sequences and extract visual features in addition to various PSE constraints. In accordance with previous methods, we evaluate our approach by matching the $200$ people in cam `\textit{a}' to $749$ people in cam `\textit{b}' and quantify the ranking quality of matchings.

\begin{figure}[t]
\centering
\includegraphics[width=0.5\textwidth]{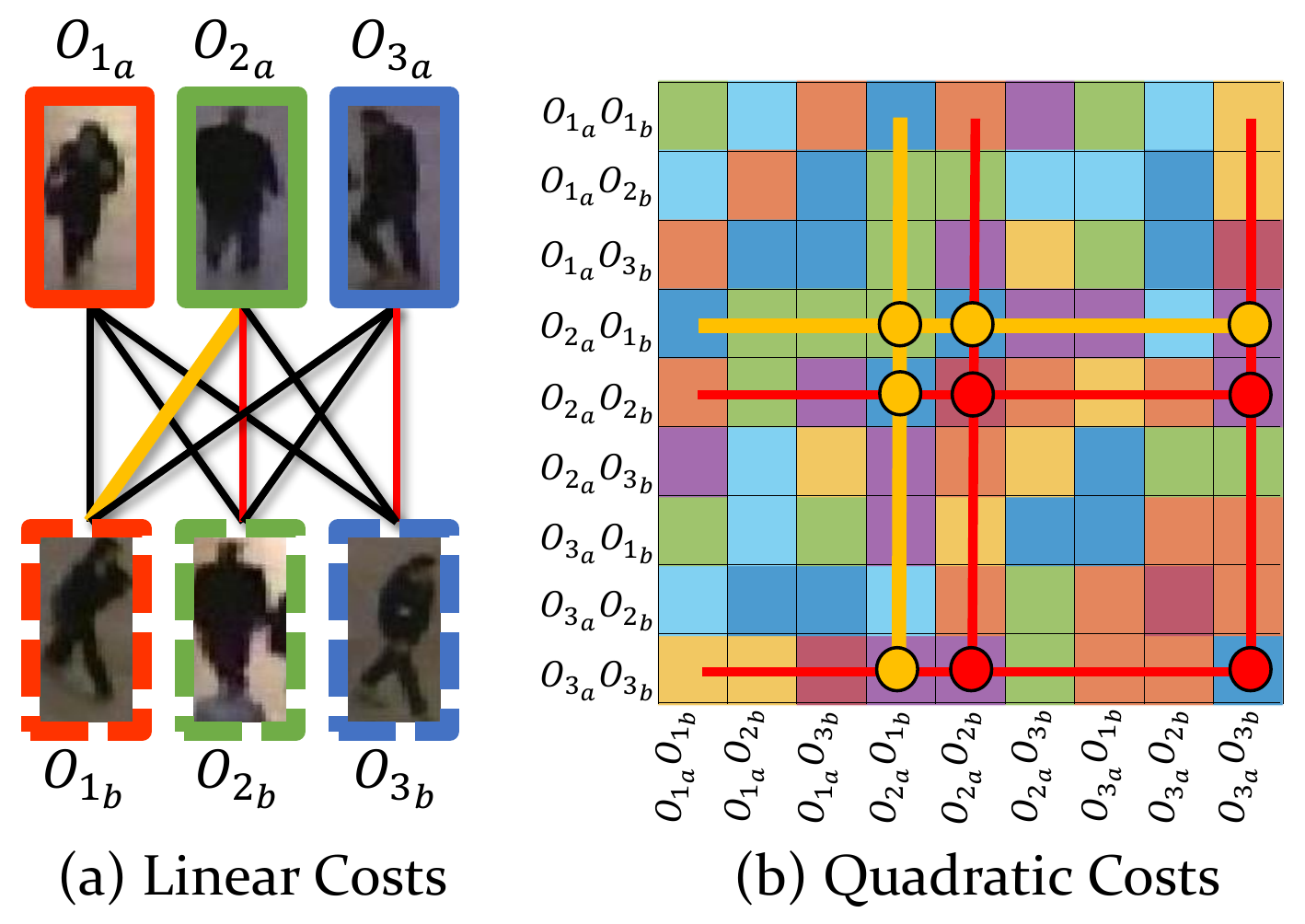}
\caption{This figure illustrates the CMC evaluation procedure with quadratic constraints. Given object tracks in the two cameras $O_{1_a}, O_{2_a}, O_{3_a}$ and $O_{1_b}, O_{2_b}, O_{3_b}$, (a) the linear constraints are computed between objects, and (b) quadratic constraints between each possible pair of matches. Adding a new match (shown with amber) requires adding one linear value and number of quadratic values equal to the size of current solution.}
\label{fig:figQuadraticExplanation}
\end{figure}

\bigskip
\noindent\textbf{Parameters:} Since there are multiple points / zones of entrances and exits, we divide the boundaries in each camera into $U_a=U_b=11$ gates. The weights used in Eq. \ref{eqL} are approximated using grid search on a separate set and then used for all the datasets. They are $\alpha_{\textrm{spt}}=\alpha_{\textrm{invColl}}=.2$, $\alpha_{\textrm{tr}}=1$, and $\alpha_{\textrm{spd}}=\alpha_{\textrm{invSpd}}=-\alpha_{\textrm{grp}}=5$. Note that, social grouping is rewarded in our formulation, i.e. people who enter and exit together in space and time are more likely to be correct matches when re-identifying people across cameras.

\begin{table*}[t]
\small
\centering
{\renewcommand{\arraystretch}{1.3}
\caption{This table presents the quantitative results of the proposed approach and other methods on the \textbf{Grand Central Dataset}. For our approach, the F-Score is reported both for Stochastic Local Search and Frank-Wolfe algorithm.}
\begin{tabular*}{1\linewidth}{@{\extracolsep{\fill}}l|c c c c c c|c}
\hline\hline
\multirow{2}{*}{\textbf{Method}}    &  \multicolumn{6}{c|}{\textbf{CMC}}                               & \textbf{F-Score}\\
\cline{2-7}
                                    &   Rank-1  & Rank-5  & Rank-10 & Rank-20 & Rank-50 & AUC (1:100) & (1-1)     \\
\hline\hline
Random                              &	 1.83\%	&  5.48\% &	 11.36\%	& 21.91\% & 54.36\%	& 51.00\% & 6.90\% \\
LOMO-XQDA \cite{liao2015person}     &	 4.06\%	& 12.37\% & 21.91\%	& 39.76\% &	71.40\%	& 63.81\% & 11.16\% \\
SDALF \cite{farenzena2010person}    &    6.09\% & 16.23\% & 23.12\% & 40.16\% & 68.56\%	& 63.01\% &	20.69\% \\
SAM	\cite{alahi2014socially}        &    6.09\%	& 27.18\% & 42.60\%	& 51.72\% &	74.44\%	& 69.60\% &	26.98\% \\
eSDC-knn \cite{zhao2013unsupervised}&   11.36\%	& 27.38\% & 38.34\%	& 50.71\% &	74.44\% & 69.49\% &	30.43\% \\
Manifold Learning (Ln) \cite{loy2013person}	            &    7.71\%	& 24.54\% &	36.71\%	& 54.97\% & 78.09\%	& 72.11\% &	 30.83\% \\
Manifold Learning (Lu) \cite{loy2013person}	            &   10.55\% & 34.08\% & 48.68\% & 66.53\% &	87.83\% & 80.50\% &	 32.66\% \\
CNN Features \cite{simonyan2014very}&   12.98\%	& 32.45\% & 44.62\% & 62.07\% & 83.77\%	& 77.79\% &	41.99\% \\
\hline
\textbf{CrowdPSE (w/o topology)}    & 	\textbf{72.62\%}	& \textbf{90.47\%} &	 \textbf{93.51\%}	& \textbf{95.74\%} &	 \textbf{97.57\%}	& \textbf{96.52\%} &	 \textbf{85.80\% / 83.59\%}\\
\textbf{CrowdPSE  (w/ topology)}   	&   \textbf{81.54\%}	& \textbf{95.33\%} & \textbf{96.15\%}	& \textbf{96.96\%} & \textbf{97.16\%}	& \textbf{96.92\%} &	 \textbf{91.54\% / 91.34\%}\\

\hline\hline
\end{tabular*}
\label{tbl:resultsGC}}
\end{table*}

\begin{table*}[t]
\small
\centering
{\renewcommand{\arraystretch}{1.3}
\caption{This table presents the quantitative results of the proposed approach and other methods on Cameras 2 and 5 of the \textbf{DukeMTMC Dataset}. We report F-score and values of Cumulative Matching Characteristic curves at ranks $1, 5, 10$, $20$ and $50$.}
\begin{tabular*}{1\linewidth}{@{\extracolsep{\fill}}l|c c c c c c|c}
\hline\hline
\multirow{2}{*}{\textbf{Method}}    &  \multicolumn{6}{c|}{\textbf{CMC}}                               & \textbf{F-Score}\\
\cline{2-7}
                                             &   Rank-1  & Rank-5  & Rank-10 & Rank-20 & Rank-50 & AUC (1:100) & (1-1)     \\
\hline\hline
Random                                       &	3.24\%	&  2.92\% &	 4.54\%	&  9.09\% & 16.88\% & 17.61\% &  0.32\%	\\
Manifold Learning (Ln) \cite{loy2013person}  & 58.44\%  & 77.60\% & 84.74\% & 88.31\% & 94.48\% & 92.06\% & 59.74\%	\\
Manifold Learning (Lu) \cite{loy2013person}  & 59.74\%	& 79.54\% & 83.77\%	& 87.66\% & 90.25\% & 88.98\% & 61.36\% \\
CNN Features \cite{simonyan2014very}         & 55.84\%	& 79.54\% & 85.71\% & 90.91\% & 95.13\% & 92.76\% & 64.94\% \\
\hline
\textbf{CrowdPSE (w/o topology)}    & 	\textbf{86.36\%}	& \textbf{97.40\%} &	 \textbf{99.35\%}	& \textbf{99.36\%} &	 \textbf{99.67\%} & \textbf{99.32\%} & \textbf{90.91\% / 90.26\%} \\
\hline\hline
\end{tabular*}
\label{tbl:resultsDUKE}}
\end{table*}

\begin{table*}[t]
\small
\centering
{\renewcommand{\arraystretch}{1.3}
\caption{This table presents the quantitative results of the proposed approach and other methods on the \textbf{PRID Dataset}. We report values of Cumulative Matching Characteristic curves at ranks $1, 5, 10$, $20$ and $50$. As can be seen, the proposed approach outperforms the existing methods.}
\begin{tabular*}{1\linewidth}{@{\extracolsep{\fill}}l|c c c c c}
\hline\hline
\multirow{2}{*}{\textbf{Method}}    &  \multicolumn{5}{c}{\textbf{CMC}}                 \\
\cline{2-6}
                                    &   Rank-1  & Rank-5  & Rank-10 & Rank-20 & Rank-50 \\
\hline\hline
KissME \cite{koestinger2012large} + Reranking \cite{leng2015person}   &	 8.00\%	& 19.00\% &	30.00\%	& 41.00\% & 57.00\%	\\
LMNN \cite{weinberger2008fast} + Reranking \cite{leng2015person}      &	 10.00\%	& 24.00\% & 34.00\%	& 44.00\% &	61.00\% \\
Mahalanobis \cite{roth2014mahalanobis} + Reranking \cite{leng2015person} & 11.00\% & 29.00\% & 37.00\% & 46.00\% & 60.00\%	\\
Non-linear ML \cite{paisitkriangkrai2015learning}                     & 17.90\%	& 39.50\% & 50.00\%	& 61.50\% &	- \\
Desc+Disc \cite{hirzer2011person}                                     & 19.18\%	& 41.44\% & 52.10\% & 66.56\% & 84.51\% \\
\hline
\textbf{CrowdPSE (w/o topology)}    & 	\textbf{21.11\%}	& \textbf{46.65\%} &	 \textbf{59.98\%}	& \textbf{76.63\%} &	 \textbf{98.81\%} \\
\hline\hline
\end{tabular*}
\label{tbl:resultsPRID}}
\end{table*}

\subsection{Evaluation Measures}

Cumulative Matching Characteristic (CMC) curves are typically used evaluating performance of re-identification methods. For each person, all the putative matches are ranked according to similarity scores, i.e. for each person $O_{i_a}$, the cost of assignment ${M_{i_a}^{j_b}} = {\langle O_{i_a}, O_{j_b} \rangle}$ is calculated for every possible match to $O_{j_b}$. Then, the accuracy over all the queries is computed for each rank. Area Under the Curve (AUC) for CMC gives a single quantified value over different ranks and an evaluation for overall performance. The advantage of CMC is that it does not require $1-1$ correspondence between matches, and is the optimal choice for evaluating different cost functions or similarity measures.

The CMC curves are meaningful only for linear constraints. Unlike linear constraints which penalize or reward matches (pair of objects), quadratic constraints penalize or reward pairs of matches. Figure \ref{fig:figQuadraticExplanation} illustrates the idea of quantifying both linear and quadratic costs through CMC, since this measure quantifies quality of costs independent of optimization. Given three objects $O_{1_a}, O_{2_a}, O_{3_a}$ and $O_{1_b}, O_{2_b}, O_{3_b}$ in cameras $a$ and $b$, respectively, the black lines in Fig. \ref{fig:figQuadraticExplanation} (a) show linear constraints~/~matchings. Let us assume we intend to evaluate quadratic constraints for the match between $O_{1_a}$ and $O_{2_b}$. For this, we assume that all other matches are correct (red lines), and proceed with adding relevant quadratic (Fig. \ref{fig:figQuadraticExplanation}) and linear costs. For evaluating match between $O_{1_a}$ and $O_{2_b}$, we add linear costs between them, as well as quadratic costs between other matches (shown with red circles in Fig. \ref{fig:figQuadraticExplanation}(b)), and pair-wise costs of the match under consideration with all other matches (shown with orange circles). This is repeated for all possible matches. Later, the matches are sorted and evaluated similar to standard CMC. Note that, this approach gives an optimization-independent method of evaluating quadratic constraints. Nonetheless, the explicit use of ground truth during evaluation of quadratic constraints makes them only comparable to other quadratic constraints.

To evaluate $1-1$ correspondence between matches, we use F-score which is defined as $2 \times (\textsf{precision} \times \textsf{recall}) / (\textsf{precision} + \textsf{recall})$ on the output of optimization. We used Hungarian Algorithm (Munkres) \cite{munkres1957algorithms} for comparison as it provides a globally optimal solution for linear costs. For the proposed PSE constraints, we use Stochastic Local Search (Sec. \ref{subsec:StochasticSearch}) and Frank-Wolfe algorithm (Sec. \ref{subsec:FW}) since we use both linear and quadratic costs.


\begin{figure}[t]
\centering
\includegraphics[width=.5\textwidth]{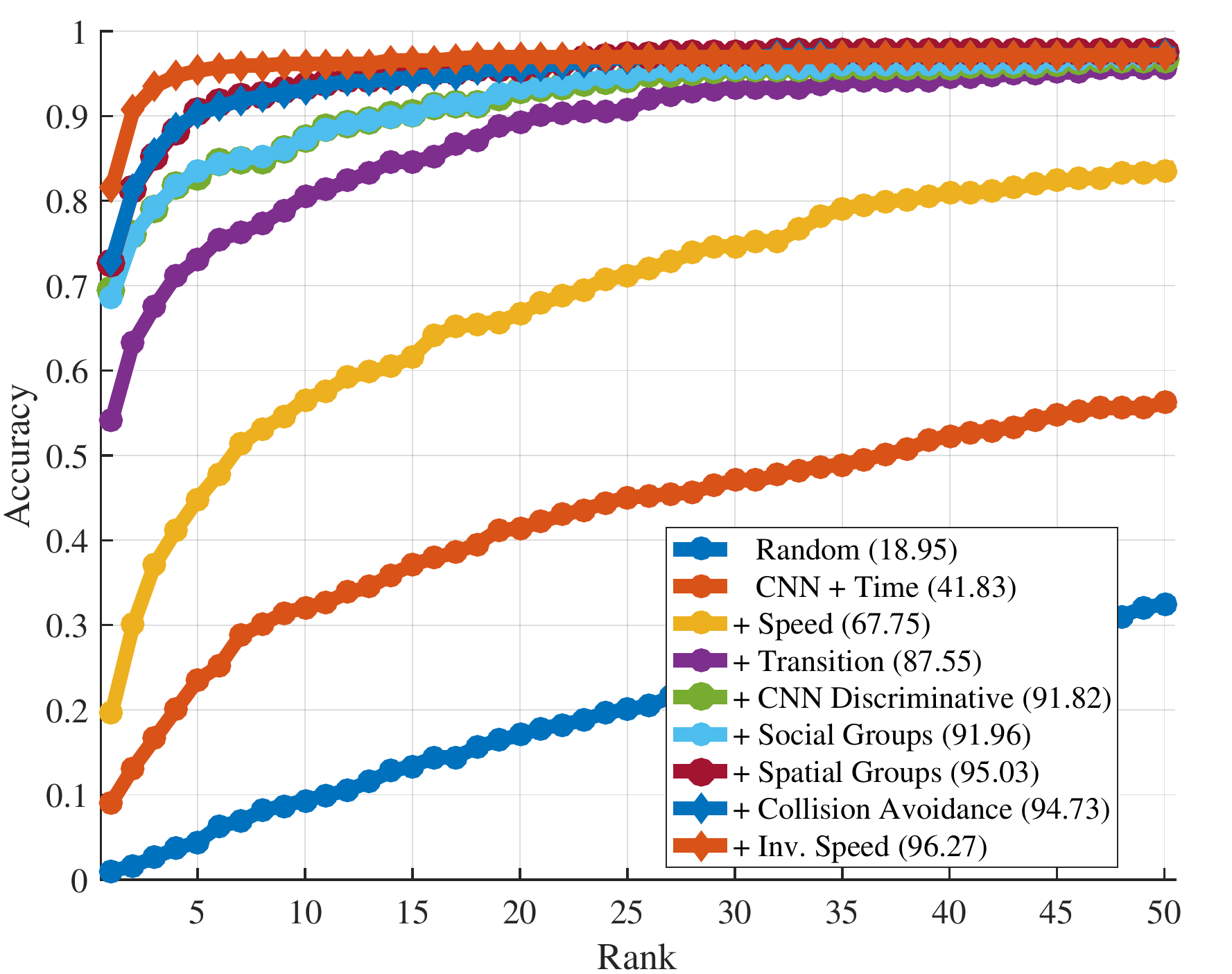}
\caption{This graph shows the CMC for different PSE constraints proposed in this paper on Grand Central Dataset. The results of random assignment are shown with blue curve, while appearance features yield the orange curve. Incorporating PSE constraints such as preferred speed (amber), transition probabilities (purple), discriminative appearance (green), social and spatial grouping (cerulean and maroon, respectively) further improve the performance. Given the topology, we can additionally incorporate collision avoidance (blue) and preferred speed in the invisible region (orange), which give the best performance. The numbers in parentheses are the AUCs between ranks 1:50 for each CMC.}
\label{fig:ranking}
\end{figure}

\subsection{Results and Comparison}


In Table \ref{tbl:resultsGC}, we present the results on Grand Central dataset of our approach with several baselines and existing methods. The first five columns show values of Cumulative Matching Characteristic curves at ranks $1, 5, 10$, $20$ and $50$. We also report Area Under the Curve (AUC) for CMC between ranks $1$ and $100$. The values of CMC are computed before any optimization. The last column shows the F-Score of $1-1$ assignments after optimization. In Table \ref{tbl:resultsGC}, the first row shows the results of random assignment, whereas next seven rows show results using existing re-identification methods. These include LOMO-XQDA~\cite{liao2015person}, SDALF~\cite{farenzena2010person}, SAM~\cite{alahi2014socially}, eSDC-knn~\cite{zhao2013unsupervised}, Manifold Learning~\cite{loy2013person}~-~normalized~(Ln) and unnormalized~(Lu), as well as CNN features~\cite{simonyan2014very} which use VGG-19 deep network. Finally, the last two rows show the results of our approach both for the case when camera topology is not known and when it is known. For 1-1 assignment, we present results for both the Stochastic Local Search as well as Frank-Wolfe algorithm in the last column. Frank-Wolfe algorithm drops the F-Score slightly compared to Stochastic Local Search. Overall, these results show that PSE constraints~-~both linear and quadratic~-~significantly improve the performance of human re-identification especially in challenging scenarios such as dense crowds.

The results on Cameras 2 and 5 of DukeMTMC dataset are shown in Table \ref{tbl:resultsDUKE}. The first row shows the results of random assignment, while results of Manifold Learning~\cite{loy2013person}~-~normalized~(Ln) and unnormalized~(Lu), as well as CNN features~\cite{simonyan2014very} are presented in the next three rows. The results from our approach are shown in the final row. Despite being an easier dataset compared to Grand Central, the PSE constraints with the proposed optimizations outperform the appearance features by a large margin. Next, we present results on PRID dataset in Table \ref{tbl:resultsPRID}. The first three rows show Reranking~\cite{leng2015person} on KissME~\cite{koestinger2012large}, LMNN~\cite{weinberger2008fast}, and Mahalanobis distance learning~\cite{roth2014mahalanobis} for re-identification. Next two rows show the performance of non-linear Metric Learning~\cite{paisitkriangkrai2015learning} and Descriptive \& Discriminative features~\cite{hirzer2011person}. The last row shows the performance of our method which is better than existing unsupervised approaches for human re-identification. For this dataset, the spatial grouping did not improve the results since the dataset captures a straight sidewalk and does not involve decision makings and different travel times between different gates. 

\subsection{Analysis}

\begin{figure}[t]
\center
\includegraphics[width=.5\textwidth]{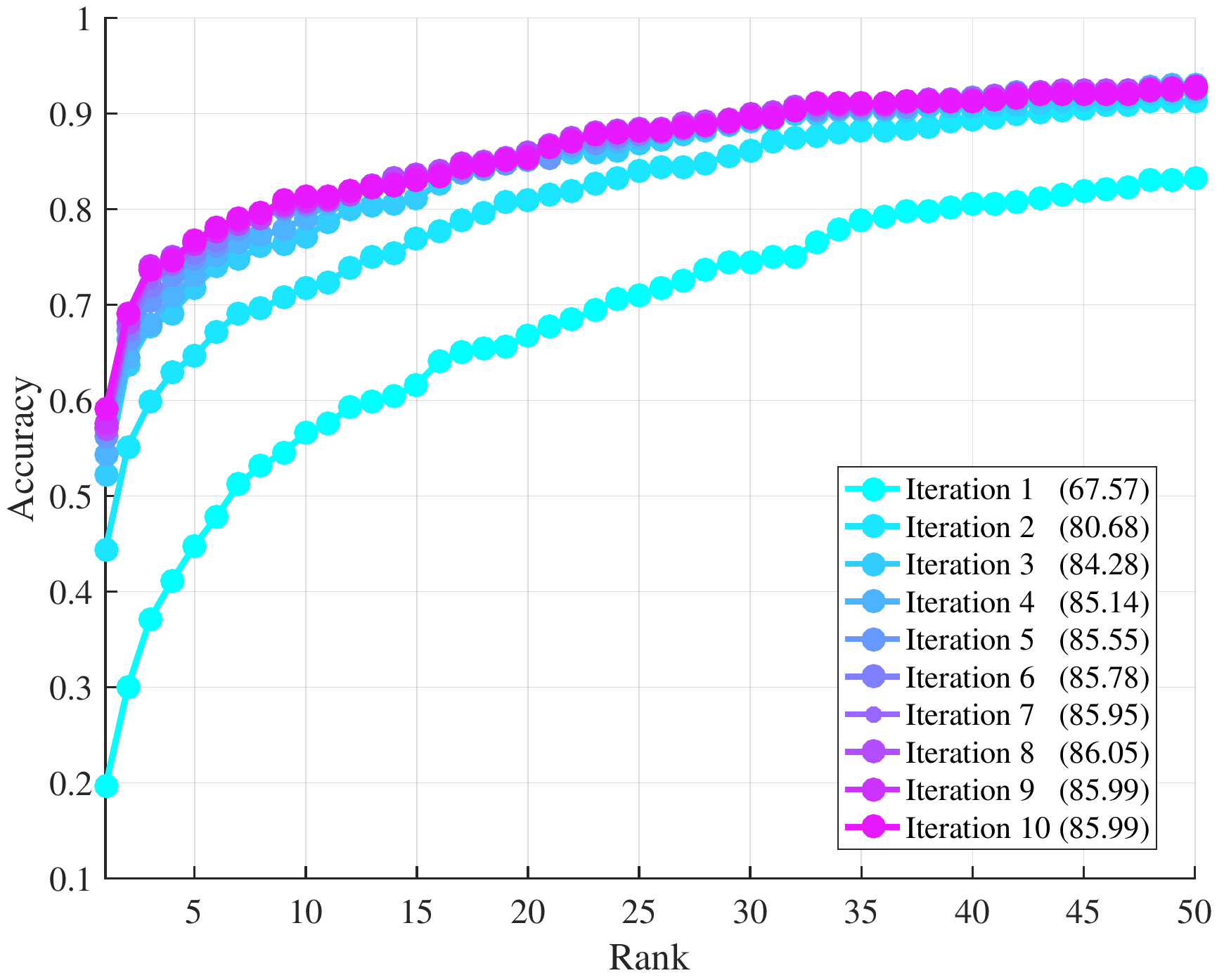}
\caption{This figure shows the improvement over iterations while learning the transition probability distributions of \emph{destination} and \emph{travel time} between gates in different cameras.}
\label{fig:iterationsPerformance}
\end{figure}

\begin{figure*}
\center
\includegraphics[width=.95\textwidth]{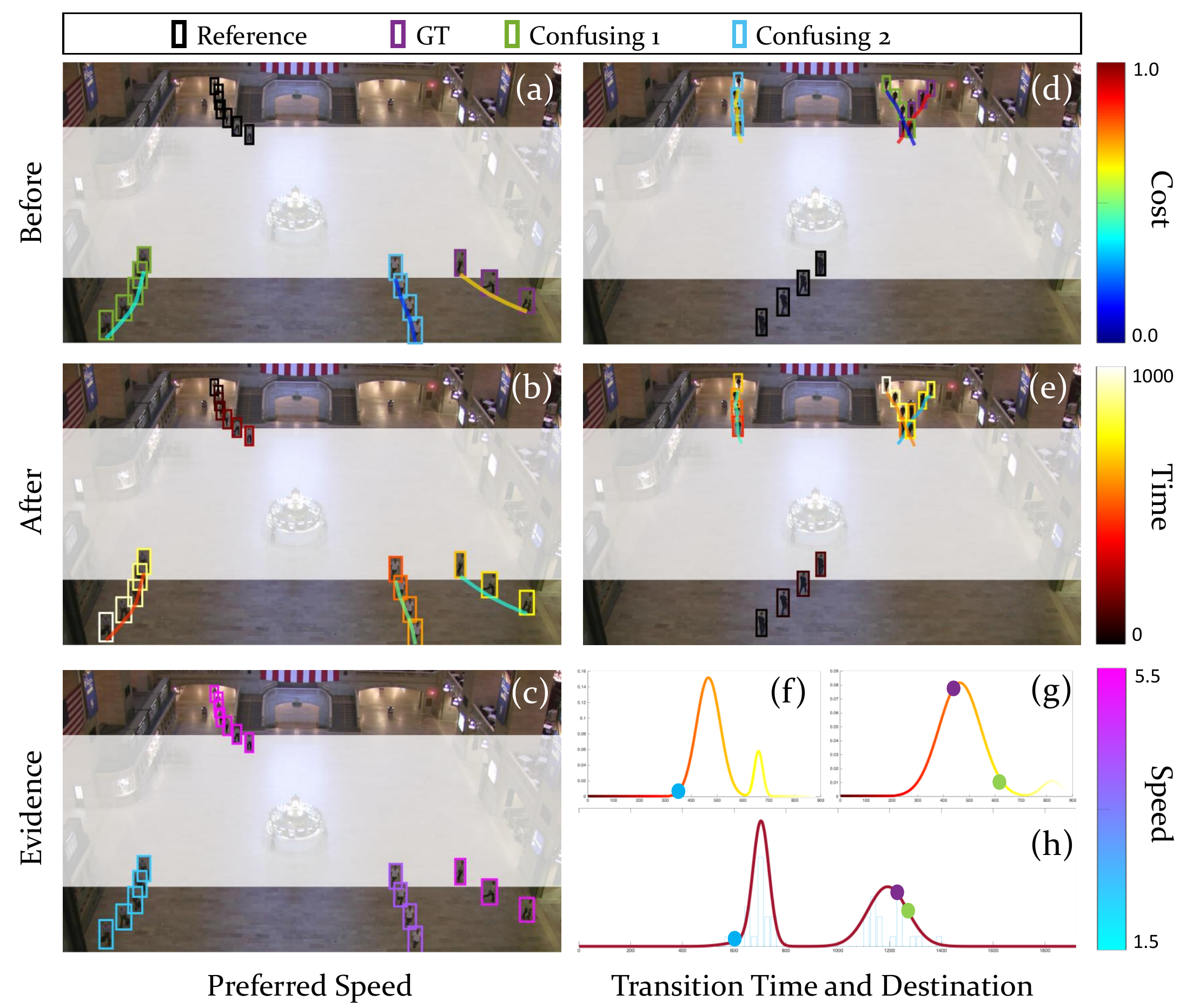}
\caption{This figure shows two real cases where \emph{preferred speed} (left) and \emph{transition time and destination} (right) improved performance of re-identification. The person under consideration is shown with black bounding boxes in first row, while the correct match is shown in purple, and two alternate hypotheses in green and blue, respectively. The tracks in first two rows are colored with respective costs shown with colorbar in the first row. In the second row, the bounding boxes depict time, colored with colorbar shown in the second row. As can be seen, the cost of the correct match reduces after application of the respective constraint. Finally, the last row presents the evidence explaining why the constraints make a difference. In (c) the bounding boxes depict preferred speeds (colored with bar on the third row), while (f,g) show the transition time probability and (h) shows the destination probability. The correct match, depicted with purple, improves most in both cases.}
\label{fig:evidenceSpeedTransition}
\end{figure*}

We performed several experiments to gauge the performance of different PSE constraints and components of the proposed approach on Grand Central dataset. The comparison of different constraints using Cumulative Matching Characteristics (CMC) is shown in Figure \ref{fig:ranking}. In this figure, the $x$-axis is the rank, while $y$-axis is accuracy with corresponding rank on $x$-axis. First, we show the results of randomly assigning objects between cameras (blue curve). Then, we use appearance features (Convolutional Neural Network) for re-identification and do not use any personal, social or environmental constraints, which we also use to compute the appearance similarity for our method  (shown with orange curve). The low performance highlights the difficult nature of this problem in crowded scenes. Next, we introduce linear constraints of preferred speed shown with amber curve which gives an improvement of $\sim26\%$ in terms of area under CMC between ranks $1$ and $50$. Similarly, transition data learned between the two cameras adds another $\sim26\%$ in terms of AUC. Figure \ref{fig:iterationsPerformance} shows the improvement of performance through  estimation of more accurate transition distributions over different iterations. Similarly, Figure \ref{fig:evidenceSpeedTransition} shows real qualitative results of preferred speed, destination and travel time, and elucidates the function of each constraint for the problem of re-identification.

\begin{figure*}
\center
\includegraphics[width=.95\textwidth]{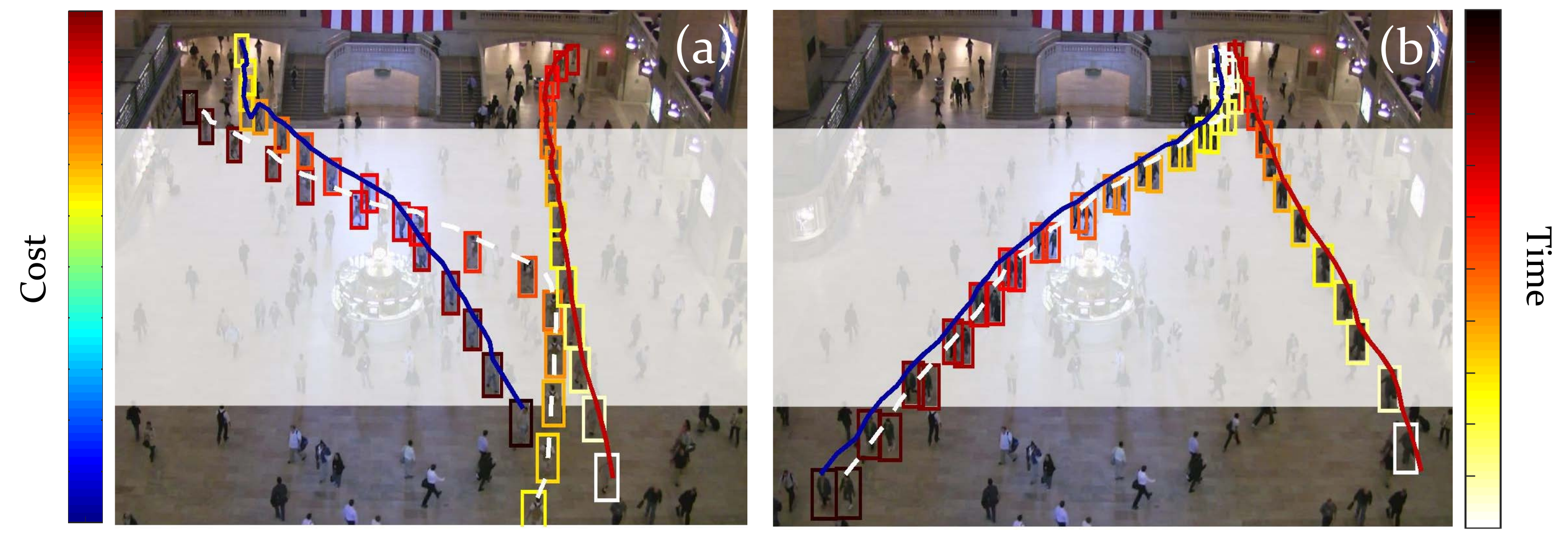}
\caption{This figure shows two examples of quadratic constraints. The color of bounding boxes indicates time using colorbar on the left, with white signifying start time and black representing end time. The person under consideration is shown with white trajectory, while the other two trajectories have the color of the cost for (a) collision avoidance and (b) grouping, color-coded with bar on the right. That is, blue and red trajectories indicate low and high costs, respectively. In (a), collision avoidance unnecessarily assigns high cost to a correct match, but not to a colliding person. On the other hand, grouping helps in re-identifying people who walk together by assigning a low cost between them.}
\label{fig:Grouping_Collision}
\end{figure*}

Next, we study the impact of quadratic constraints of social (cerulean curve) and spatial (maroon curve) grouping, both of which make slight improvement to the matching performance, with combined effect of $\sim3\%$ improvement in Rank-1 and $\sim8\%$ in Rank 5 performance. Note that both these quadratic constraints are antipodal in the sense that former rewards while latter penalizes the loss function. The last two curves in Figure \ref{fig:ranking} show the performance using constraints computable if camera topology is known. Given topology, we employ collision avoidance shown in blue (diamond markers), whereas the constraint capturing the desire of people to walk with preferred speed between cameras is shown in orange (diamond markers), which gives the maximum AUC of $96.27\%$ in conjunction with other PSE constraints. Beyond $90\%$, the increments in AUC appears small, however a noticeable improvement in performance is visible between Ranks 1:10 which is crucial for 1-1 assignment.



This study shows that except for collision avoidance, all PSE constraints contribute significantly to the performance of human re-identification. We provide real examples of collision avoidance and social grouping in Fig. \ref{fig:Grouping_Collision}(a) and (b), respectively. In Fig. \ref{fig:Grouping_Collision}, the bounding boxes are color-coded with time using colormap shown on left. White-to-Yellow indicate earlier time stamps while Red-to-Black indicate later ones. The person under consideration is shown with dashed white line, while the track of two other people in each image are color-coded with costs using colormap on the right. Here, blue indicates low cost whereas red means high cost.

Collision avoidance which has been shown to work for tracking in non-crowded scenes \cite{pellegrini2009you} deteriorates the results slightly in crowded scenes. Fig. \ref{fig:Grouping_Collision}(a) shows a case where collision avoidance constraint assigns a high cost to a pair of correct matches. Due to limitation in space in dense crowds, people do not change their path significantly. Furthermore, any slight change in path between cameras is unlikely to have any effect on matching for re-identification. On the other hand, the grouping constraint yields a noticeable increase in performance as evident in Fig. \ref{fig:Grouping_Collision}(b) This is despite the fact that the Grand Central dataset depicts dense crowd of commuters in a busy subway station, many of whom walk alone.

\section{Conclusion} \label{secConclusion}

This paper addresses the problem of re-identifying people across non-overlapping cameras in crowded scenes. Due to the difficult nature of the problem, the appearance similarity alone gives poor performance. We employ several personal, social and environmental constraints in the form of \textit{appearance, preferred speed, destination, travel time},
and \textit{spatial and social grouping}. These constraints do not require knowledge about camera topology, however if available, it can be incorporated into our formulation. Since the problem with PSE constraints is NP-hard, we proposed stochastic local search, and the computationally efficient Frank-Wolfe algorithm that can handle both quadratic and linear constraints. The crowd dataset used in the paper brings to light the difficulty and challenges of re-identifying and associating people across cameras in crowds, while the personal, social and environmental constraints highlight the importance and utility of extra-appearance information available in videos for the task of re-identification.

\bigskip
\noindent\textbf{Acknowledgment:} This material is based upon work supported in part by, the U.S. Army Research Laboratory, the U.S. Army Research Office under contract/grant number W911NF-14-1-0294.

%

\bibliographystyle{IEEETran} 
\bibliography{PSECrowds}


\begin{IEEEbiography}[{\includegraphics[width=1in,height=1.25in,clip,keepaspectratio]{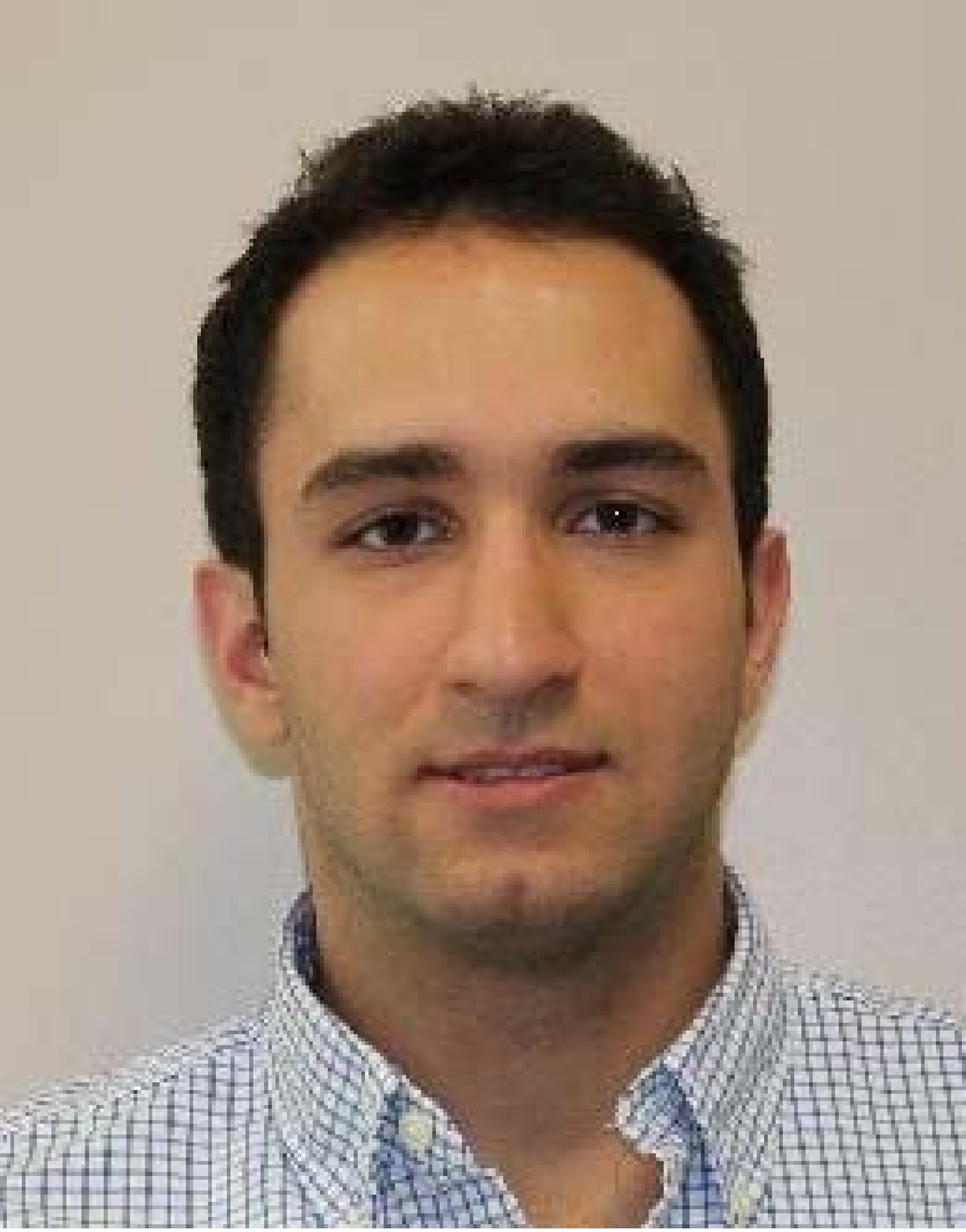}}]{Shayan Modiri Assari}
received his B.S. degree in Electrical Engineering from Sharif University of Technology in 2011 and the M.S. degree in Computer Engineering from University of Central Florida in 2015. He is currently a PhD candidate in University of Central Florida's Center for Research in Computer Vision (CRCV). He has published papers in European Conference on Computer Vision, Computer Vision and Pattern Recognition, and Journal of Machine Vision and Applications. His research interests include event detection, action recognition, object tracking, person re-identification, video surveillance and graph theory.
\end{IEEEbiography}

\begin{IEEEbiography}[{\includegraphics[width=1in,height=1.25in,clip,keepaspectratio]{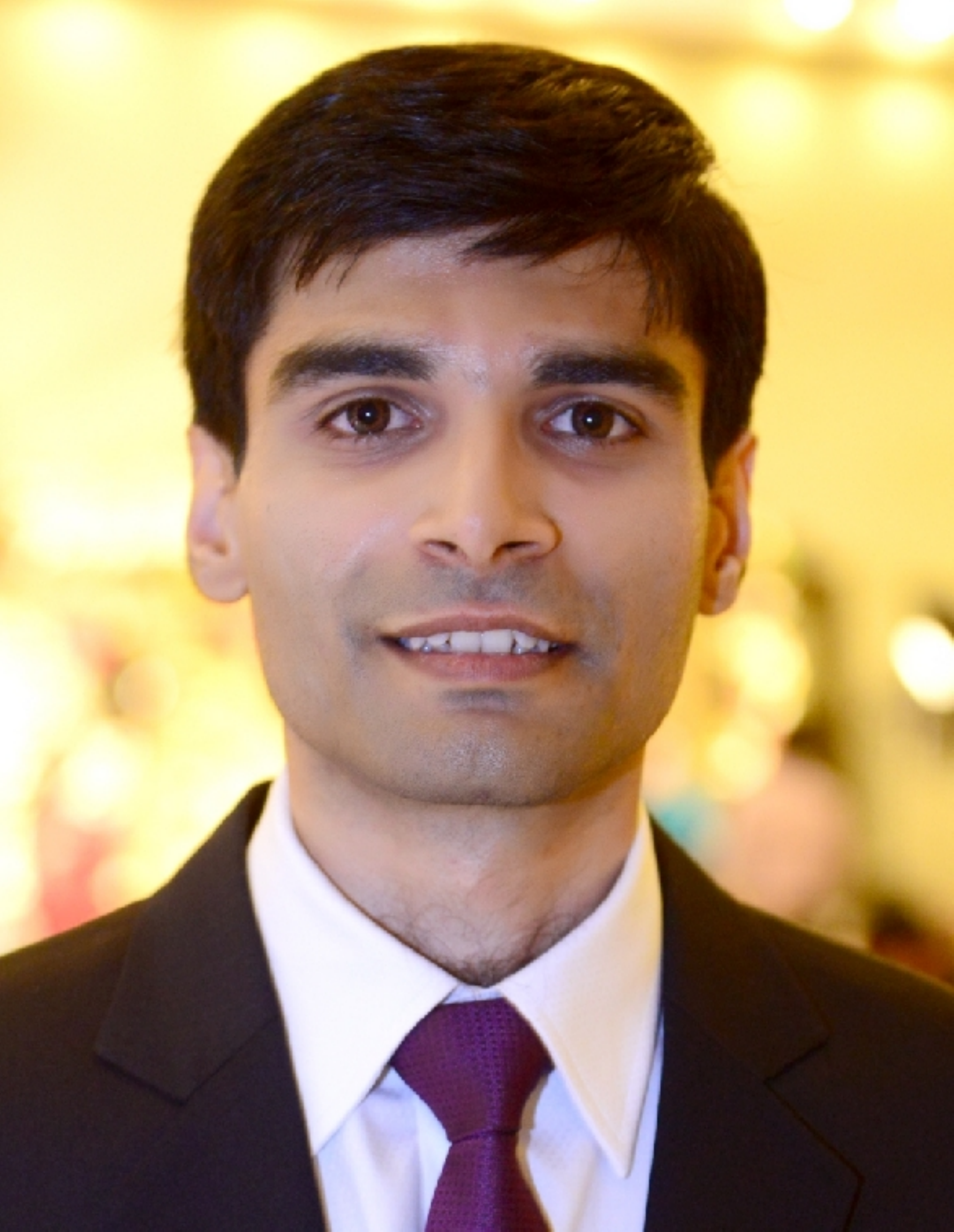}}]{Haroon Idrees}
is a Postdoctoral Associate at the Center for Research in Computer Vision at University of Central Florida. He has published several papers in conferences and journals such as CVPR, ICCV, ECCV, Journal of Image and Vision Computing, Computer Vision and Image Understanding, and IEEE Transactions on Pattern Analysis and Machine Intelligence. His research interests include crowd analysis, person detection and re-identification, visual tracking, multi-camera and airborne surveillance, action recognition and localization, and multimedia content analysis. He received the BSc (Hons) degree in Computer Engineering from the Lahore University of Management Sciences, Lahore, Pakistan in 2007, and the PhD degree in Computer Science from the University of Central Florida in 2014.
\end{IEEEbiography}

\begin{IEEEbiography}[{\includegraphics[width=1in,height=1.25in,clip,keepaspectratio]{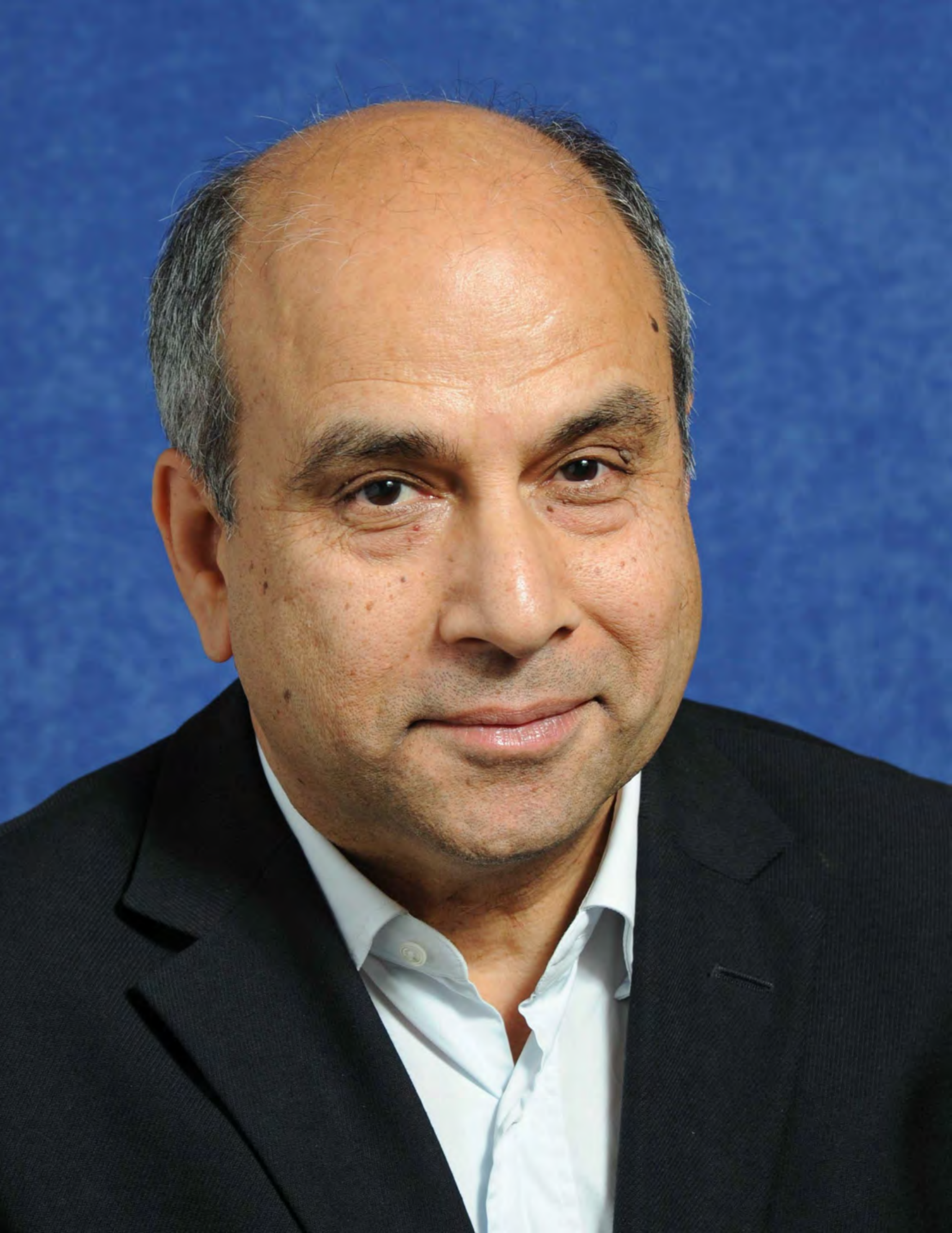}}]{Mubarak Shah,}
the Trustee chair professor of computer science, is the founding director of the Center for Research in Computer Vision at the University of Central Florida (UCF). He is an editor of an international book series on video computing, editor-in-chief of Machine Vision and Applications journal, and an associate editor of ACM Computing Surveys journal. He was the program cochair of the IEEE Conference on Computer Vision and Pattern Recognition (CVPR) in 2008, an associate editor of the IEEE Transactions on Pattern Analysis and Machine Intelligence, and a guest editor of the special issue of the International Journal of Computer Vision on Video Computing. His research interests include video surveillance, visual tracking, human activity recognition, visual analysis of crowded scenes, video registration, UAV video analysis, and so on. He is an ACM distinguished speaker. He was an IEEE distinguished visitor speaker for 1997-2000 and received the IEEE Outstanding Engineering Educator Award in 1997. In 2006, he was awarded a Pegasus Professor Award, the highest award at UCF. He received the Harris Corporation�s Engineering Achievement Award in 1999, TOKTEN awards from UNDP in 1995, 1997, and 2000, Teaching Incentive Program Award in 1995 and 2003, Research Incentive Award in 2003 and 2009, Millionaire�s Club Awards in 2005 and 2006, University Distinguished Researcher Award in 2007, Honorable mention for the ICCV 2005 Where Am I? Challenge Problem, and was nominated for the Best Paper Award at the ACM Multimedia Conference in 2005. He is a fellow of the IEEE, AAAS, IAPR, and SPIE.
\end{IEEEbiography}

\end{document}